\documentclass{article}
\pdfoutput=1

\usepackage[preprint]{neurips_data_2022}

\usepackage[utf8]{inputenc} 
\usepackage[T1]{fontenc}    
\usepackage{url}            
\usepackage{booktabs}       
\usepackage{amsfonts}       
\usepackage{nicefrac}       
\usepackage{microtype}      
\usepackage{comment}
\usepackage{hyperref}       
\usepackage{xcolor}  
\usepackage[pdftex]{graphicx}
\usepackage{float}
\usepackage{verbatim}
\usepackage{babel}
\usepackage{enumitem}
\usepackage{multicol}
\usepackage{ragged2e}
\usepackage{graphicx}
\usepackage{subcaption}

\begin{document}

\title{The Canadian Cropland Dataset: A New Land Cover Dataset for Multitemporal Deep Learning Classification in Agriculture}

%

\author{%
    Amanda A. Boatswain Jacques\textsuperscript{\rm 1,2}, Abdoulaye Baniré Diallo\textsuperscript{\rm 1}, Etienne Lord\textsuperscript{\rm 2} \\
    \textsuperscript{\rm 1}
    Département d’informatique, Laboratoire d’algèbre, \\ 
    de combinatoire et d’informatique mathématique (LACIM) \\
    Université du Quebec à Montréal \\
    PO BOX 8888 Downtown Station, H3C 3P8 \\
    \textsuperscript{\rm 2}
    Agriculture and Agri-Food Canada \\
    430 Gouin Boulevard,  Saint-Jean-sur-Richelieu, J3B 3E6 \\
    \texttt{\url{boatswain_jacques.amanda@courrier.uqam.ca}} \\
}

\maketitle
\begin{abstract} 
 Monitoring land cover using remote sensing is vital for studying environmental changes and ensuring global food security through crop yield forecasting. Specifically, multitemporal remote sensing imagery provides relevant information about the dynamics of a scene, which has proven to lead to better land cover classification results. Nevertheless, few studies have benefited such high spatial and temporal resolution data due to the difficulty of accessing reliable, fine-grained and high-quality annotated samples to support their hypotheses. Therefore, we introduce a temporal patch-based dataset of Canadian croplands, enriched with labels retrieved from the \textit{Canadian Annual Crop Inventory}. The dataset contains 78,536 manually verified and curated high-resolution (10 m/pixel, 640 x 640 m) geo-referenced images from 10 crop classes collected over four crop production years (2017-2020) and five months (June-October). Each instance contains 12 spectral bands, an RGB image, and additional vegetation index bands. Individually, each category contains at least 4,800 images. Moreover, as a benchmark, we provide models and source code that allow a user to predict the crop class using a single image (ResNet, DenseNet, EfficientNet) or a sequence of images (LRCN, 3D-CNN) from the same location. In perspective, we expect this evolving dataset to propel the creation of robust agro-environmental models that can accelerate the comprehension of complex agricultural regions by providing accurate and continuous monitoring of land cover.\\
  

\end{abstract}

\section{Introduction}

The term \textit{land cover}, defined as the surface cover on the ground, comprises anything including vegetation, urban development, mountains or others. Whereas land cover corresponds to a physical property of the ground, \textit{land use} refers to the function or purpose attributed to a piece of land. These functions fall within various categories such as recreation, agriculture, lodging, business or governmental. The ability to accurately map, delineate and characterize land cover provides essential information which can improve resource management, activity planning and change detection \cite{NRSC}. Monitoring the land cover and land use changes in perennial, annual and cover crops can help agronomists and agricultural agencies to improve management and address issues related to climate change, global food security and biodiversity \cite{MAZZIA2020, Khatami2016}. This is done through the analyses of remote sensing (RS) data obtained via instruments such as satellites in orbit, or with proximal sensing using unmanned aerial vehicles (UAV) like drones, which generate data with high spatial resolution \cite{Ghamisi2019, Zhang2016, Zhu2017}. A recent surge of open-access remote sensing (RS) image data has led to an incredible increase in research opportunities, while simultaneously empowering agronomists with data-driven tools \cite{Bovolo2015, Kumar2018}.

\begin{figure}[t!]
    \centering
    \includegraphics[width=0.75\textwidth]{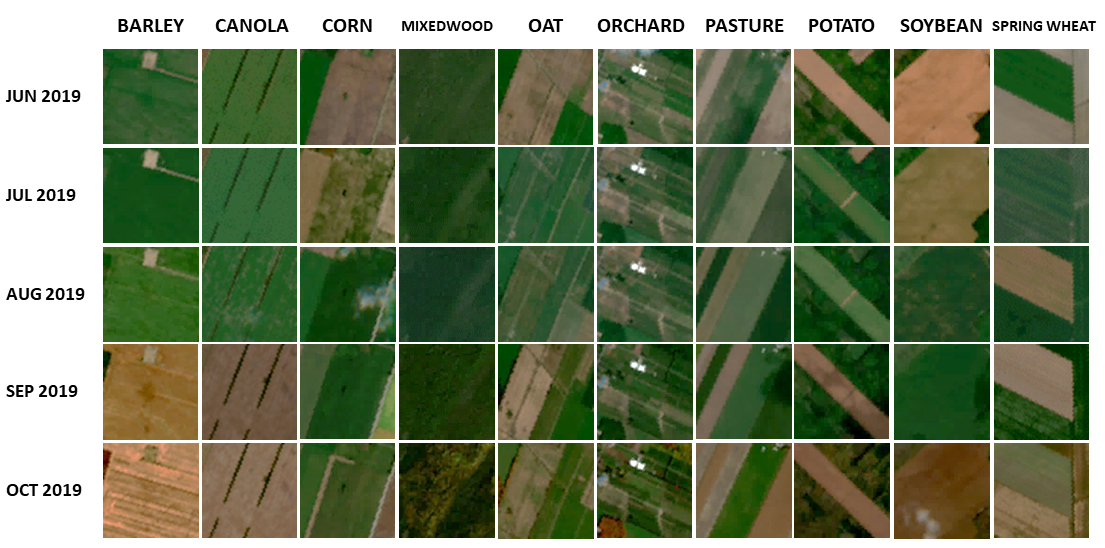}
    \caption{An image mosaic of RGB images taken from the 2019 data of the \textit{Canadian Cropland Dataset}. The dataset contains Sentinel-2 images of North American crops over 5 distinct seasonal periods (June to October). The dimensions of each image are 640 m $ \times $ 640 m (64 $ \times $ 64 pixels). } \label{fig:can_crop_temporal_mosaic}
\end{figure}

\paragraph{land cover classification} 
For decades, RS has played an integral role in the development of precision agriculture applications \cite{Ghamisi2019, Yang2018}. One of these applications is land cover classification, which attempts to distinguish between different types of land cover classes present in a scene. For example, many studies \cite{Wessel2018,KAMILARIS2018, Kumar2018, Isaac2017, TEIMOURI2019, Kussul2017} have presented models that can classify and segment land cover  with surprisingly high accuracy. However, challenges continue to persist in agricultural applications such as the high intra-class variability of fields, the lack of high quality data devoid of artifacts such as cloud cover, landscape complexity and the variability of scale \cite{Ghamisi2019, Khatami2016}. This can be mitigated with the acquisition of high quality labeled RS data through human visual observation, but this process remains extremely time-consuming due to the diversity in crop geometry and phenology, as well as differences in overall composition exhibited in agricultural croplands \cite{Ghamisi2019,zhao2021evaluation}. For this reason, the majority of studies \cite{Khatami2016, Kussul2017, campos2020understanding, Masoud2020, Song2019} attempting pixel-wise or patch-wise classification of RS imagery are done either with a single reference image taken mid-season or analyze a small study area \cite{Sumbul2021} which makes it difficult to exploit the benefits of deep learning (DL) architectures. 

Significant advances in deep learning (DL) are still required to develop models capable of tackling real RS problems at a global scale with high reliability and repeatability \cite{Helber2019, Ghamisi2019, Sumbul2021}. We provide a valuable resource to help close this gap by presenting the \textit{Canadian Cropland Dataset} (Figure \ref{fig:can_crop_temporal_mosaic}). In sum, \textbf{our contributions} are the following: (1) a collection of image patches of Canadian agricultural croplands containing over 75,000 unique instances retrieved from the \textit{Sentinel-2} satellite constellation, (2) a RS dataset that can be used in both temporal studies and in fixed-time analyses, and finally (3) an evaluation of the performance of agricultural land classification using a dynamic \textit{Long-Term Recurrent Convolutional Network} (LRCN) and 3D-CNN network architectures against more traditional static image analysis architectures such as the ResNet, DenseNet and EfficientNet.



 Supervised classification of land cover is one of the most prominent active research areas in RS hyperspectral image analysis \cite{Zhu2017, Ghamisi2019, WANG2018, Sumbul2021}. Early pioneering studies have explored the use of handcrafted features constructed using algorithms to differentiate between scenes and land cover types \cite{JIA2013, BENEDIKTSSON2003, Zhang2016}, while others have incorporated texture information and/or ancillary data \cite{Khatami2016, Ghamisi2019}. More recent studies have explored  DL architectures in higher level (i.e. pixel-level) classification of land cover \cite{Li2018} and the automatic delineation of land cover boundaries (\textit{i.e.} semantic segmentation) \cite{Masoud2020}. Deep Neural Networks (DNN) infer high-level hierarchical features that describe the complex non-linear relationships that exist between the spectral information of satellite imagery and the observed material or scene \cite{Ghamisi2019}. When sufficient data is available, these networks can perform well even in the presence of noisy data \cite{Jung2021}. Several examples of DL applications include work by Song \textit{et al.} \cite{Song2019}, who used a Convolutional Neural Network (CNN) to classify land cover with medium-resolution (30 m) imagery from the \textit{Landsat-8} satellite. This CNN outperformed other methods such as an SVM and an RF, but had a low average overall accuracy falling below 67\% for all models. Another work by Masoud \textit{et al.} \cite{Masoud2020} developed a Multiple-Dilation Fully Convolutional Network (MD-FCN) to classify agricultural field boundaries in imagery retrieved from Sentinel-2. The FCN performed semantic segmentation (pixel-wise image classification) while the addition of dilated kernels preserved resolution without adding parameters to be learned. Although lower resolution boundaries were fragmented or missing, general boundaries were properly labeled.
 
The emergence of modern satellite sensors, such as Sentinel-2, has generated a wealth of rich spectral data with both high temporal resolution (a few days between measurements) and high spatial resolution (from 10 m/pixel) at a low cost. Consequently, the temporal parameter can now be exploited as a new dimension in analyses to compare pairs of images or time series when performing land cover classification \cite{Ghamisi2019,Bovolo2015,Khatami2016}. Several examples of studies using multitemporal data include work by Mazzia \textit{et al.} \cite{MAZZIA2020}, who developed a pixel-recurrent CNN to classify pixels within 15 types of agricultural crops. Their proposed model exploited traits such as time correlation, temporal pattern extraction and multi-class classification modules, and provided very high classification accuracies (above 90\%) compared to other baseline models. Kussul \textit{et al.} \cite{Kussul2017} developed a 2D-CNN to perform patch-based classification (7 $ \times $ 7 pixels) of land cover in Ukraine using a combined dataset of Landsat-8 and Sentinel-1 imagery at four different time periods in the growth season. The 2D-CNN differentiated between 11 classes (8 corresponding to major agricultural crops) with very high accuracy (94.6\%). More recently, Campos-Taberner \textit{et al.} \cite{campos2020understanding} investigated the use of a Bi-Long Short-Term Memory (Bi-LSTM) network using pixel-based Normalized Difference Vegetation Index or NDVI (see Appendix \ref{definition_vegetation_indices}: \textit{Definition of Vegetation Indices}) data provided by Sentinel-2 to follow the phenology and classification of 16 major crop types in Spain. They obtained high-precision levels between 92.9\% to 99.9\%. In another study \cite{zhao2021evaluation}, a pixel-based 1D-CNN, an LSTM, a Gated-Recurrent Units (GRU), an LSTM-CNN, and a GRU-CNN model were used to classify 7 crop types in China with incomplete Sentinel-2 imagery (only 10 bands). Even with ~43.5\% missing data, their classifiers achieved an accuracy between 81.21\% and 86.57\%. However, despite those significant results and high performance reported in these experiments, no datasets were provided to extend their results to other regions. 

 Some noticeable RS datasets closest to our work are the recently created \textit{BigEarthNet-MM} \cite{Sumbul2021}, \textit{Eurosat} \cite{Helber2019}, \textit{CropHarvest} \cite{tseng2021cropharvest}, \textit{Calcrop21} \cite{Ghosh2021CalCROP21AG} and \textit{Denethor} \cite{kondmann2021denethor} datasets. The \textit{BigEarthNet-MM} \cite{Sumbul2021} is a multimodal, multi-label dataset containing 590,326 pairs of Sentinel-1 and Sentinel-2 image patches collected since June 2017. It currently features 12 Sentinel-2 spectral bands as TIFF images with dimensions of 120 $ \times $ 120 pixels (1,200 $\times$ 1,200 m, for the 10 m bands) collected in 10 European countries (Austria, Belgium, Finland, Ireland, Kosovo, Lithuania, Luxembourg, Portugal, Serbia, Switzerland). This dataset was manually validated through visual inspection and was annotated using the CORINE land cover map of 2018 with 43 classes of land cover, including agricultural classes such as rice fields, vineyards, fruit trees and berry plantations, olive groves, pastures, annual crops associated with permanent crop, complex cultivation patterns, and agro-forestry areas.   Alternatively, the \textit{Eurosat} dataset contains a total of 27,000 land use images in 10 classes and is also noteworthy for its high-spectral resolution (12 bands per image, Sentinel-2) but lacks in the granularity of the crop classes (annual crop, permanent crop, pasture). Although some of these multitemporal datasets (\cite{tseng2021cropharvest, kondmann2021denethor,Ghosh2021CalCROP21AG}) are similar in the number of major crop categories (at least 9), spatial and spectral resolution, our dataset is unique due to its large geographical area and high spatio-temporal information (more than 2 years) on top of the Sentinel bands. Moreover, ~70\% of geolocations have an image collection of at least 10 images (over 4 years), allowing a temporally rich monitoring of a field through a multi-year, multi-crop rotation. 

\section{Dataset creation} 
The \textit{Canadian Cropland Dataset} can be directly downloaded from our \href{https://github.com/bioinfoUQAM/Canadian-cropland-dataset}{github repository}\footnote{https://github.com/bioinfoUQAM/Canadian-cropland-dataset}, along with the code and models presented in this paper. The dataset is released under the \textit{Montreal Data License} (see Appendix \ref{dataset_license}) and we povide a specific \textit{Datasheet for Datasets} (see Appendix \ref{canadian_crop_datasheet}).

\paragraph{Collection of satellite imagery}\label{collection_satellite_imagery} 
The images in the dataset originate from the \textit{Sentinel-2} constellation, which comprises two satellites (Sentinel-2A and 2B) that share the same sun-synchronous orbit phased at 180 degrees to each other. On average, the constellation takes a complete image of the Earth every 5 days, creating a collection of roughly 18 registered images per month for the region of Canada \cite{SENTINEL2}. Google Earth Engine (GEE) was selected as the main image manipulation platform due to its easy access to large records of satellite data, ancillary data, processing algorithms and free cloud computing resources and \cite{GORELICK2017, Kumar2018}. Images with a cloud cover percentage above 5\% were rejected, and we applied a final bitwise operation on the "cloud mask" band Q60 to remove any remnant cloud or shadow pixels. Following this, a median image for each month was produced by calculating the median of the values across all pixels and spectral bands in a collection. Median images of croplands at five different periods in the growth season (June, July, August, September and October) were retrieved. Each image is composed of 12 spectral bands which range from the Visible (VNIR) and Near Infra-Red (NIR) to the Short Wave Infra-Red (SWIR) wavelengths \cite{SENTINEL2}. All the bands and their descriptions are listed in Table \ref{sentinel_bands} of the Appendix section. Finally, a region of $ 64 \times 64 $ pixels centered on each field between the years 2017 to 2020 was extracted and downloaded directly from GEE, giving rise to a rich multi-year dataset with unique long-term crop rotation patterns.

\begin{figure}[t]
    \centering
    \includegraphics[width=0.80\textwidth]{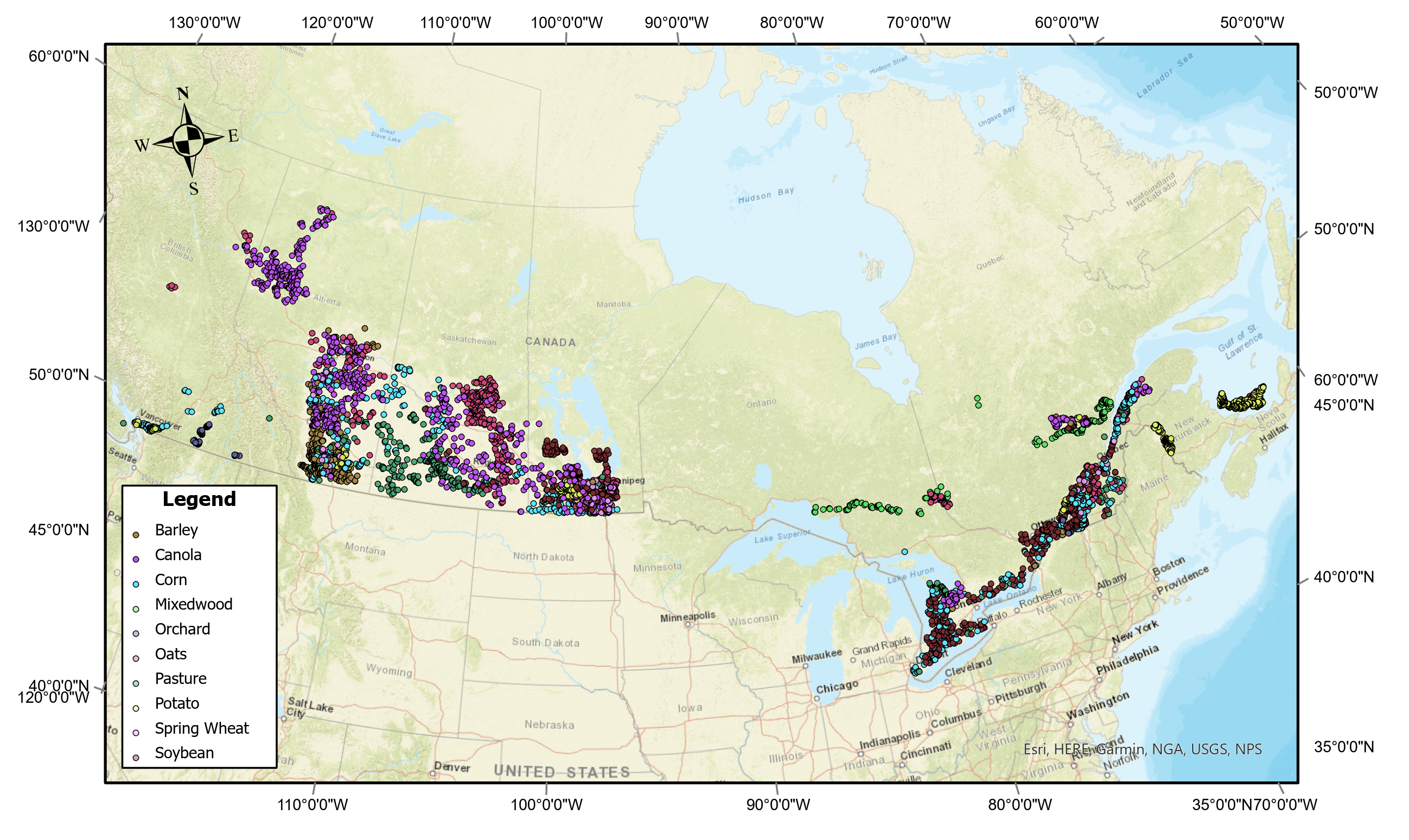}
    \caption{Map representing an overview of the selected geographical locations used in the \textit{Canadian Cropland Dataset}. Markers are randomly chosen fields and are color-coded by the 2019 crop types. The map was created using ArcGis Pro 2.8.0 (\url{https://esri.com/en-us/argis/}).} \label{fig:ACI_crop_inventory}
\end{figure}

\paragraph{Selection of geographical points}
To label each image, we used the \textit{Canadian Annual Crop Inventory} (ACI) developed by Agriculture and Agri-Food Canada (AAFC) and distributed using an \textit{Open Government License}. The overall accuracy of the crop inventory is at least 90.56\% at a final spatial resolution of 30 m. Table \ref{tab:ACI_accuracy_tab} (Appendix \ref{ACI_accuracy_table}) provides a summary of the regional crop class accuracies. Locations were selected randomly from 10 predominant North American crops with layers available in the ACI: barley (\textit{Hordeum vulgare}), canola (\textit{Brassica napus}), corn (\textit{Zea mays}), mixedwood, oat (\textit{Avena sativa}), orchard crops, pasture, potatoes(\textit{Solanum tuberosum}), soybeans (\textit{Glycine max}(L.) Merr.) and spring wheat (\textit{Triticum aestivum} (L.). A total of 6,633 geographical points were manually validated by overlapping the 2019 crop type mask with each Sentinel-2 median image. Figure \ref{fig:ACI_crop_inventory} shows the entire study region considered, as well as the classes of the selected points. 

\paragraph{Creation of vegetation indices} 
Vegetation indices quantify traits such as the amount of biomass, the vigor or the growth dynamics present in a spectral image \cite{Wessel2018, Xue2017}. Using the original Sentinel-2 bands, we derived additional bands for 5 common vegetation indices (see Appendix \ref{definition_vegetation_indices} for mathematical definitions). These image transformations were applied directly on the GEE cloud computing platform. After gathering the additional vegetation index bands, a composite image was generated by duplicating the information across 3 channels and saving them as .png files for ease of use with deep learning frameworks. For example, the created images are compatible with a ResNet architecture. An RGB dataset was also created by combining the visible red, green and blue bands (identified as the Sentinel-2 bands: B4, B3, B2) in a single image (see Figure \ref{barley_example} in Appendix).

\paragraph{Data cleaning} 
The data were curated using visual observation to exclude any low quality image that contained satellite artifacts or cloud pixels missed by the previously applied filters (see section \ref{collection_satellite_imagery}:\textit{ Collection of satellite imagery}). The cleaning process was done using the RGB images and the tagging of improper images was performed by 2 trained individuals. For each geographical location, the raw data is supplied to the user in the form of a .zip file with all the original Sentinel-2 bands and composite bands. 

\paragraph{Training, Validation and Test sets} 
The dataset comes prepackaged in dedicated training (70\%), validation (15\%) and testing sets (15\%) suitable for use with Keras or Tensorflow image data generators. These sets remain the same for each individual year and image type (i.e NDVI, PSRI, etc.). Images belonging to the same geographical coordinates were kept together during the dataset splitting process to reduce overfitting. 
\section{Dataset analysis}

\begin{figure}[t]
    \centering
    \includegraphics[width=0.60\textwidth]{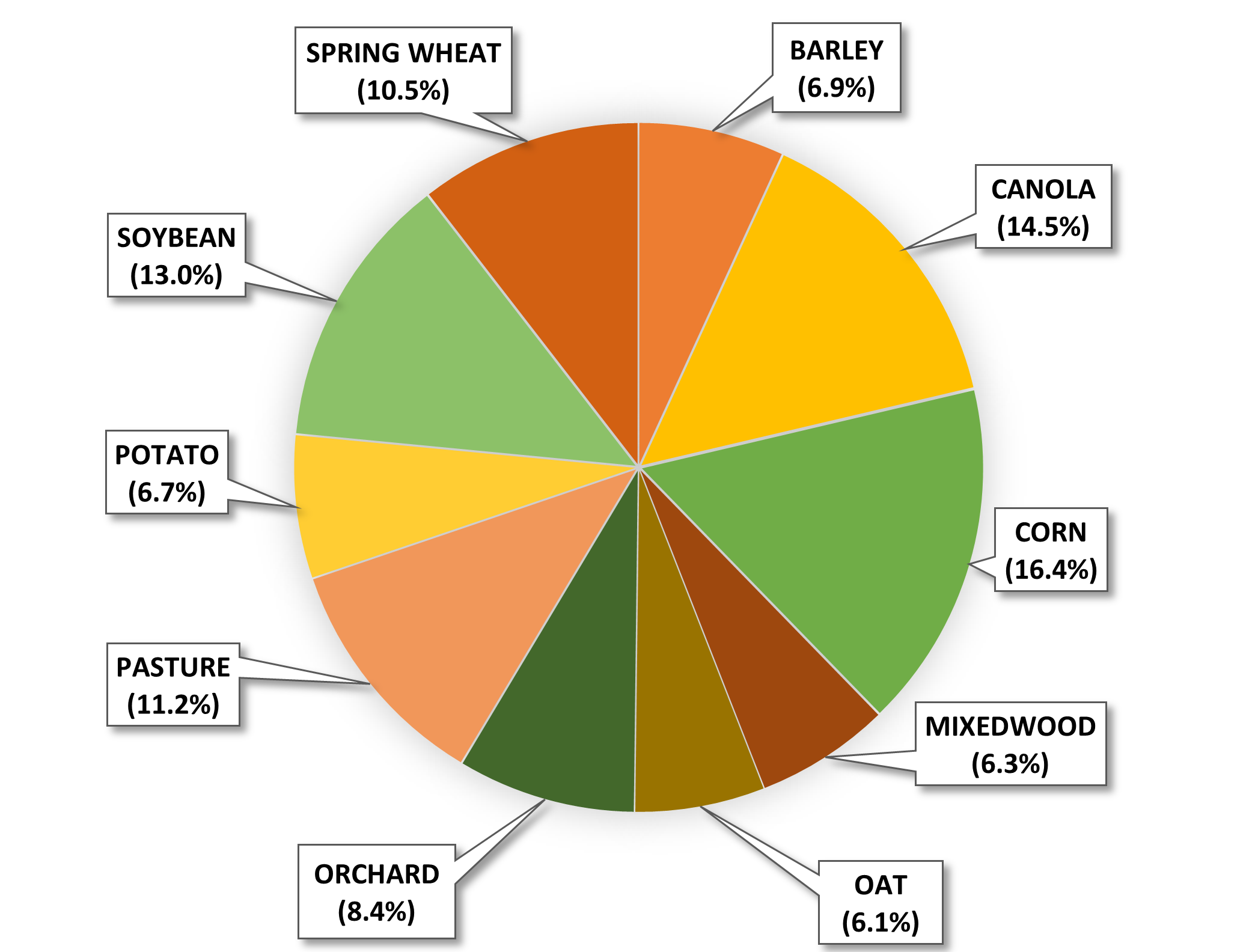}
    \caption{A pie chart showing the distribution of 10 major crop categories from the collected images for 2017 to 2020. Since different crop rotations happened during this period, multiple classes and crop patterns can be observed at a single location over time.  } \label{fig:dataset_distribution}
\end{figure}

\paragraph{Dataset image distribution} 
Analyses were performed to calculate the number of images originating from each time period (year, month), crop types and provinces. A total of 46 crop classes were initially observed from the crop rotation patterns. The final version of the dataset targets the 10 most common crops that were present in Canada over the study period (Figure \ref{fig:dataset_distribution}). The final image counts for each category are: barley: 5,382; canola: 11,366; corn: 12,878; mixedwood: 4,981; oats: 4,807; orchards: 6,594; pasture and forages: 8,797; potatoes: 5,294; soybean: 10,208; spring wheat: 8,229. Additional figures and full statistics on the dataset can be consulted in the supplementary material. Roughly 88.2 \% of all points have 3 or more images available for each growth season. However, looking at the monthly analyses, images from the month of June were less common overall. This may be due to the high percentage of cloud cover that is associated with the spring season in Canada and that was observed within the dataset (\textit{data not shown}). To mitigate this, missing data could be retrieved using a multi-source data fusion approach by gathering images from another available satellite at the same resolution (e.g. Landsat-8) or UAVs. Finally, for our crops of interest, most images came from the provinces of Quebec (26.46\%), Alberta (16.34\%), Manitoba (13.80\%), Saskatchewan (12.39\%) and British Colombia (10.92\%). This may be due to some crop types being more prominent in different climatic zones, such as mixedwood which is highly concentrated in the southern regions of Quebec and Ontario, or barley fields being mostly located in the provinces of Alberta, Saskatchewan and Manitoba.

\begin{figure}[t]
    \centering
    \includegraphics[width=0.65\textwidth]{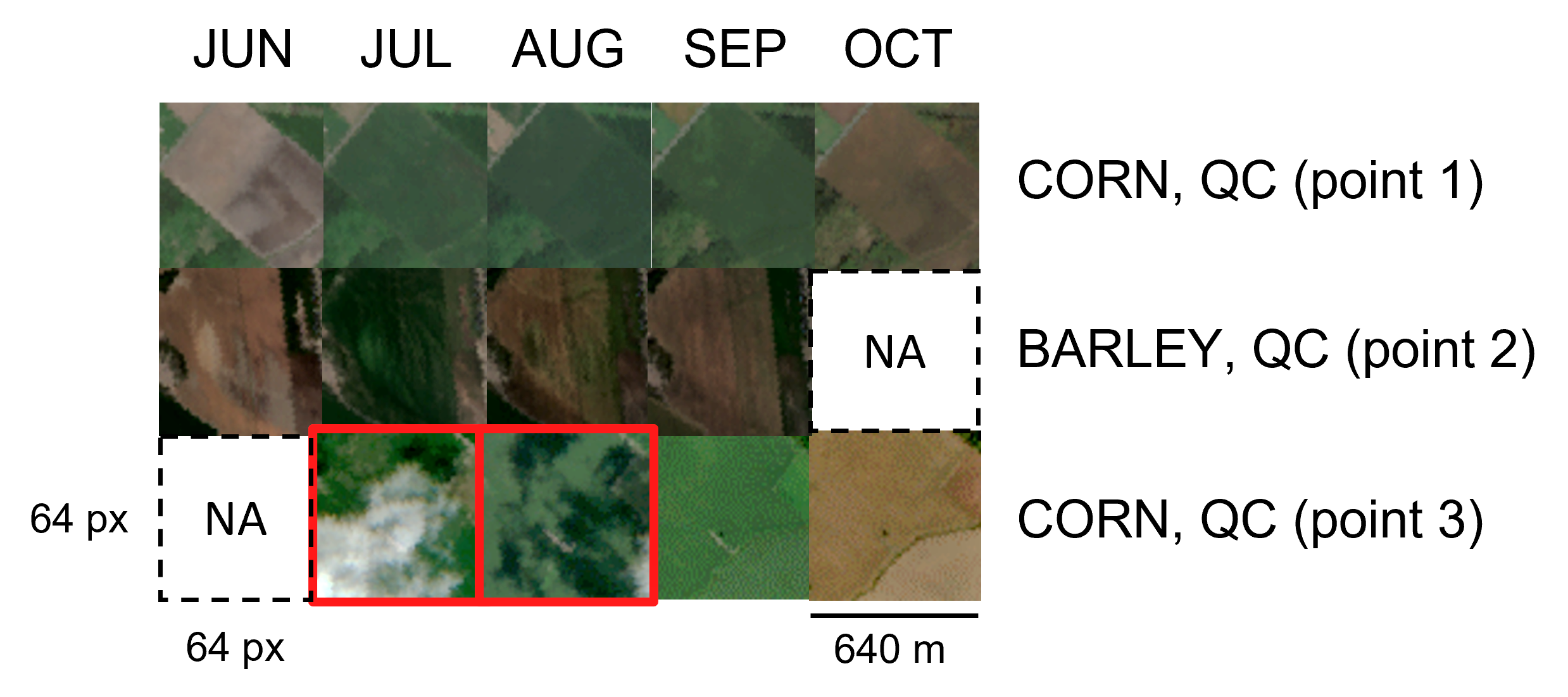}
    \caption{Examples of collected images for three locations (point 1 to 3). At the right side of the figure, supplementary information are shown including the crop class and province acronym. An example of removed images due to the presence of clouds are indicated by a red outline and missing images by \textit{NA}. The image dimension is 64 pixels (px) corresponding to 640 m for each side. } \label{fig:missing_images}
\end{figure}

From the 6,633 geographical points selected, a total of 93,175 images were successfully retrieved, which is less than the 132,660 ($ 5 \times 6333 \times 4$) temporal samples that were theoretically expected over the four year period. This was primarily due to the occasional lack of available Sentinel-2 data when the image either had too much cloud cover or when the image quality was too low. An example of this scenario is depicted in Figure \ref{fig:missing_images}. The top row (point 1) shows a location where imagery from all 5 time steps was available. The middle (point 2) and bottom (point 3) rows illustrate both the cases of either missing (NA) or low-quality data (red border). 

\section{Benchmarking experiments}  
In this section, we present two classification experiments performed using the 2019 version of the dataset. In the first experiment, we treated each individual image as a training instance (static image classification), similar to the work that was performed in \cite{Helber2019}. In the second experiment, we explored the use of a temporal image series as an input. Materials, methods and final results are described in the sections below. The models were trained either on a Dual Intel Xeon Gold 5120 workstation with 64 Gb of RAM and a Titan RTX graphics card (24 Gb of RAM) or Quadro RTX 8000 (48 Gb or RAM). Different deep learning models developed in the Python (version 3.8.8) programming language were trained using the Keras (version 2.4.3) and TensorFlow (version >=2.1.0) frameworks. Each model was trained three times (N=3) with a different random seed. We report the average of these 3 test results using performance metrics defined in Appendix \ref{performance_metrics}. 
 
\subsection{Static image classification}

\paragraph{ResNet-50, DenseNet, EfficientNet} 
Different ResNet-50 \cite{He2016} architectures were employed to perform land cover classification. We first used a ResNet-50 pre-trained with ImageNet to reduce overfitting of the network. In this case, the base ResNet-50 architecture was supplemented with a dense layer (1024 neurons, ReLu activation) followed by a dropout layer (25\%), as well as a final prediction layer (10 neurons, softmax activation). This network was trained according to the following specifications: 30 epochs (10 with the base layers untrainable), a batch-size of 128, an rmsprop optimizer and a learning rate set at 0.001. We also trained some ResNet backbones from scratch by first using the keras-tuner (version 1.0.1) \cite{omalley2019kerastuner} HyperResNet model to estimate the best hyperparameters using the RGB images, including the number of 3- and 4- convolutional blocks, the type of pooling layers (maximum or average), the optimizer and the learning rate for 10 epochs using RandomSearch. Finally, the best parameters for the untrained ResNet were a total depth of 4 for the 3-Conv layers and 23 for the 4-Conv layers, an svg optimizer, an average pooling and a learning rate of 0.01 which was used for those experiments. More recent network architectures such as DenseNet-121 \cite{Huang2016DenseNet} and EfficientNet-B0 \cite{Tan2019EfficientNet} were also evaluated. In both cases, we used the Keras implementations pre-trained with ImageNet, selecting the more light-weighted architecture to avoid over-fitting with our small sized images. In order to be comparable to the ResNet architectures, we added the same top prediction layers (dense layer of 1024 neurons with ReLu, 25\% dropout and softmax activation) with the exception that a BatchNormalization layer was added after the GlobalAveragePooling2D layers of the EfficienNet-B0. Training was performed using the same specifications as the ResNets, but with only the top 20 layers of the EfficienNet being trainable (with the exception of the BatchNormalization layers set as untrainable). In total, the dataset used for training, validation and testing of the static image classification represented 22,413 distinct images (Table \ref{tab:training_counts}).   

\subsection{Dynamic image classification}
\paragraph{LRCN and 3D-CNN trained from scratch} 
For the temporal analysis, we implemented a \textit{Long-Term Recurrent Convolutional Network} (LRCN) model described by \cite{Donahue2017} and a 3D-CNN inspired by \cite{Zunair20203dcnn} using the Keras deep learning framework. The LRCN network has a CNN as a hierarchical and spatial feature extractor coupled with a Long Short-Term Memory (LSTM) model to recognize the sequential patterns in the spatio-temporal data. The CNN network was built with 2 ConvLSTM \cite{SHI2015} (kernel size = (7,7) in layer 1, and (3,3) in layer 2) and Max Pooling blocks, with each one followed by a Dropout layer (50\% block 1, 50\% block 2) and ReLu activation. We determined the number of layers for the CNN portion of the network empirically. The output was then reshaped before being passed to an LSTM layer, a Dropout layer (50\%) and a final prediction layer (10 neurons, softmax activation). We tested different combinations of hidden units in the last LSTM layer, which were identified as one of the most influential model hyperparameters in \citep{Donahue2017} (see Table \ref{tab: LRCN_perf_metrics} in the Appendix \ref{LRCN_perf_metrics}). The best overall combination was to use 64 hidden units. Finally, this network architecture was trained according to the following specifications: 30 epochs, a batch-size of 32, an rmsprop optimizer and a learning rate set at 0.001. For the 3D-CNN, the network was built using a series of 3 Conv3D layers with 32 filters, each with kernel size of (1,2,2) followed by BatchNormalization. We dropped the MaxPooling3D layers of the original implementation \cite{Zunair20203dcnn} since our temporal series was limited in length. These blocks were followed by a GlobalAverage3D pooling layer, a dense layer with 512 neurons with ReLu activation, a 50\% dropout layer and a final prediction layer (10 neurons, softmax activation). The training parameters for the 3D-CNN resembled those of other models except for the number of epochs which was reduced to 15. For the dynamic experiments, training instances were sequences or "image sets" of 3 consecutive images from 2019 from the same data point. In contrast to the static experiment, the dataset used for training, validation and testing represented only 7,502 distinct instances since some locations did not present sequences of 3 consecutive valid images (Table \ref{tab:training_counts}). 

\begin{table}[ht]
    \caption{Number of training, validation and tests samples per experiment (RGB 2019 dataset)} \label{tab:training_counts} 
        \centering
        \begin{tabular}{l c c}
            \toprule
                                 \multicolumn{3}{c}{Experiment type}        \\
            \cmidrule(r){2-3}
                             &  Static image classification   &   Dynamic image classification     \\
            \hline
                 Training    &      15,612                    &   5,210            \\
                 Validation  &      3,349                     &   1,100             \\
                 Testing     &      3,452                     &   1,192             \\
                 Total       &      22,413                    &   7,502             \\ 
            \bottomrule
        \end{tabular}
      
\end{table}

\subsection{Benchmarking}

\paragraph{Results}
A summary of our results are presented in Figure \ref{fig:classification_results} and complete results are reported in Table \ref{tab:Model_perf_metrics} of Appendix \ref{Model_perf_metrics}. For the first static experiment, where we use the pre-trained ResNet-50, the training conditions were similar to those in Helber \textit{et al.} \cite{Helber2019}. Here, we found that the highest accuracy was obtained with the RGB images (0.67) followed by OSAVI (0.62), NDVI45 (0.55), GNDVI (0.52), NDVI (0.52) and PSRI (0.48).  Other models behaved similarly \textit{i.e.} models trained on RGB images performed better than those trained on a single vegetation index which was also observed in another study \cite{Helber2019}. Those results are, however, lower than some of the accuracies reported by other authors for datasets created with Sentinel-2 imagery \cite{campos2020understanding, zhao2021evaluation} with higher than 85\% accuracy, and higher than 98\% for land cover classification \cite{Helber2019}. The ResNet architecture selected with the keras-tuner and also obtained much lower overall accuracy. 
Although it was the most modern network, the DenseNet \cite{Huang2016DenseNet} resulted in the lowest classification accuracy (0.40), while the EfficientNet-B0 \cite{Tan2019EfficientNet}, with only 5.38 Millions parameters (vs 25.7 Millions for ResNet-50 and 8.1 Millions for DenseNet-121) resulted in a classification accuracy of 0.53 for RGB images. Training the models for a higher number of epochs did not improve results. Overall, discrepancies between the classification accuracies were expected since, by design, some of the crops are not as predominant in the 2019 dataset (e.g. oats, Figure \ref{fig:dataset_distribution}). Moreover, the agricultural landscape of Canada displays variable topologies, field shapes and sizes which are reflected by discrepancies in management practices (e.g. center pivot irrigation in the Canadian prairies) \cite{Amani2020}. 

In the second experiment, we investigated the possibility of improving agricultural land use classification using a temporal series of 3 consequent images. The LRCN trained with most vegetation indices (GNDVI, NDVI, NDVI45 and OSAVI) showed low average precision, ranging from 0.03 to 0.14, and low accuracy from 0.27 to 0.30  (see Table \ref{Model_perf_metrics} in the Appendix). However, training using the PSRI vegetation index resulted in both an average precision (0.45) and accuracy (0.56) higher than the pre-trained ResNet-50. For the 3D-CNN, the average accuracy were higher when trained with vegetation indices including PSRI, resulting in values between 0.28 and 0.57, with average precision still below 0.50. Training the LRCN with RGB images maximized average accuracy (0.77, Figure \ref{fig:classification_results}), representing a ~10\% higher score than the pre-trained ResNet50 with RGB. Also, the LRCN classification showed strong average precision (0.61), recall (0.64) and F1-score (0.62). The 3D-CNN, with only 31,082 trainable parameters (vs 14.6 Millions for the LRCN), resulted in an unexpected higher average accuracy of 0.77, with similar average precision (0.61), recall (0.62) and F1-score (0.60). Those results are similar to the 77\% accuracy obtained by Amani \textit{et al.} \cite{Amani2020} using Sentinel-1, -2 images, and an ANN on the GEE platform. In their case, they used NDVI and the Normalized Difference Water Index (NDWI) as their features derived from Sentinel-2 data. They also used 197,634 images to represent their 17 crop classes (including corn, soybean and potato). Different reasons can explain the poor performance obtained by our models trained with a singular vegetation index. For NDVI and related indices \cite{Delegido2011, GITELSON1996}, a saturation of the index occurs for different cultures when they attain maturity (end of vegetative stage). In contrast, the PSRI index measures the senescence or ripening of the crop cover, and thus, might present more learnable features in the months of July, August and September, more present in our dataset \cite{Merzlyak1999}.

When comparing both of the dynamic classification models, the 3D-CNN, although having a simplistic and un-optimized architecture in our implementation, showed similar results to the more complex LRCN architecture composed of ConvLSTM layers. Similar results were obtained in \cite{OKSUZ2019136} when analyzing temporal cardiac magnetic resonance images, with results favoring either model depending on the prior data augmentation techniques. Thus, more research is required to improve these network architectures and optimize them for use with this new dataset. Nevertheless, the difference in the number of parameters is $\sim$ 450 $ \times $ more for the LRCN network compared to the 3D-CNN, which leads us to prefer the latter architecture since the gains in accuracy are limited when comparing model size. However, the similar results obtained by the two network architectures might also be due to some limitations in the number of training samples which could be solved though data augmentation as demonstrated in \cite{OKSUZ2019136}. One limitation arising from the current benchmark is that we did not evaluate classification of the same location (i.e static vs dynamic) since misclassification could happen in one static image through the season, thus needing a rule-based or ensemble algorithm to correctly identify the resulting class in order to compare a period of 3 months. 


\begin{figure}[ht]
    \centering
    \includegraphics[width=1.0\textwidth]{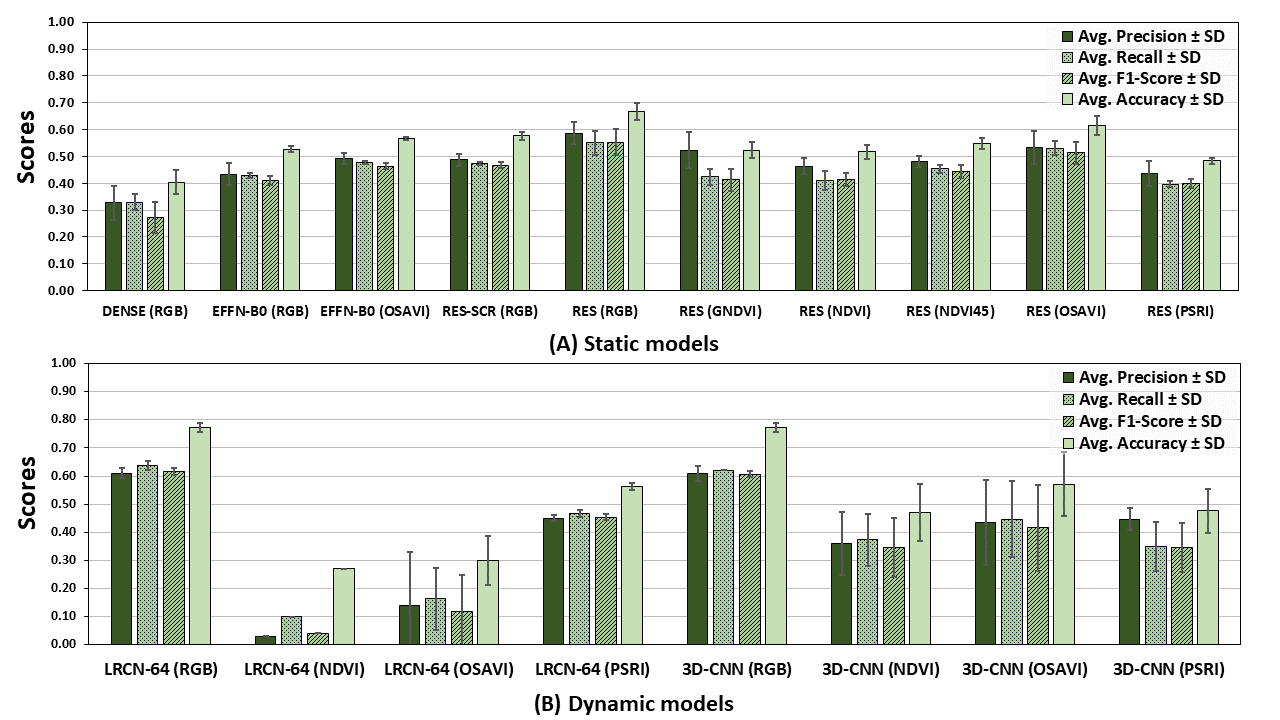}
    \caption{Summary of classification results for models trained on the 2019 subsets. In A, results for the static image classification and in B, results for the dynamic image classification. Full results are presented in Table \ref{tab:Model_perf_metrics} in the Appendix. \textit{Abbreviations:} DENSE (DenseNet-121 pre-trained with ImageNet), EFFN-B0 (EfficientNet-B0 pre-trained with ImageNet), RES (ResNet-50 pre-trained with ImageNet), RES-SCR (ResNet-50 trained from scratch). Results reported are average ± standard deviation (SD) of triplicate experiments with different starting random seeds.} \label{fig:classification_results}
\end{figure}

\section{Contributions and Future Work}

The availability of annotated multitemporal images which can accurately represent, according to \cite{Ghamisi2019}, \textit{i}) all multitemporal classes, \textit{ii}) the inter-relation between classes along the time-series with a reliable statistic, and \textit{iii}) the high temporal and spatial variability in large scenes is scarce. The \textit{Canadian Cropland Dataset} bridges this gap by directly addressing points \textit{i}) and \textit{iii}) in an agricultural context. One potential negative societal impact could be the development of models allowing the advance monitoring of crop conditions and agricultural fraud detection. However, one positive impact could be the classification of land use based on different crop rotations. Crop rotation is becoming central to the study of higher land resilience, carbon sequestration and biodiversity \cite{li2021crop}. Thus, this dataset might be used to develop new models fostering large-scale studies of biological conservation measures. Furthermore, different opportunities exist to augment the current dataset using the provided scripts. First, the USDA CropScape - Cropland Data Layer \cite{USDACropland2011} is also available on GEE. Incorporating more field samples from the data layer could allow a greater representation of both the diversity of North American croplands and their topology. Second, a European cropland layer known as the \textit{Denmark LPIS dataset} was used by \cite{dKerselaersLIPS2019} to compare the loss of agricultural land, and could also be used to increase the dataset scope beyond North America. Finally, ancillary data such as meteorological data could be added using the location information. This additional information could help to better understand vegetation phenology correlated with the included vegetation indices layers.     

\section{Conclusion}
The release of Google Earth Engine and Sentinel-2 imagery to the RS research community have catalyzed the creation of new tools, models and data sources for land cover and land use classification. We exploited these tools by creating a multitemporal patch-based dataset of Canadian cropland images and identified the challenges that arise when curating a large dataset using unprocessed satellite imagery. Futhermore, we performed benchmarking tests on a subset of the dataset using two different strategies: static (e.g. ResNet, DenseNet, EfficientNet) or dynamic series of images (e.g. 3D-CNN, LRCN). These tests demonstrated that incorporating spatio-temporal sequences resulted in better land cover classification accuracy using our dataset, while reducing the number of required training instances and training parameters. We believe that this novel dataset can be used in the creation of robust environmental models that can help to better understand and map complex and dynamic agricultural regions. 

\paragraph{Acknowledgments}
This work was supported by Agriculture and Agri-Food Canada. All authors contributed equally to the production of this manuscript.

\bibliographystyle{plain}
\bibliography{neurips_submission.bib}

\begin{thebibliography}{10}

\bibitem{ACI}
{Agriculture and Agri-Food Canada}.
\newblock Annual crop inventory.
\newblock \url{https://open.canada.ca/en/apps/aafc-crop-inventory}, 2016.

\bibitem{Amani2020}
Meisam Amani, Mohammad Kakooei, Armin Moghimi, Arsalan Ghorbanian, Babak
  Ranjgar, Sahel Mahdavi, Andrew Davidson, Thierry Fisette, Patrick Rollin,
  Brian Brisco, and Ali Mohammadzadeh.
\newblock Application of google earth engine cloud computing platform, sentinel
  imagery, and neural networks for crop mapping in canada.
\newblock {\em Remote Sensing}, 12:3561, 11 2020.

\bibitem{BENEDIKTSSON2003}
J.A. Benediktsson, M.~Pesaresi, and K.~Amason.
\newblock Classification and feature extraction for remote sensing images from
  urban areas based on morphological transformations.
\newblock {\em IEEE Transactions on Geoscience and Remote Sensing},
  41(9):1940--1949, Sep. 2003.

\bibitem{USDACropland2011}
Claire Boryan, Zhengwei Yang, Rick Mueller, and Mike Craig.
\newblock Monitoring us agriculture: the us department of agriculture, national
  agricultural statistics service, cropland data layer program.
\newblock {\em Geocarto International}, 26(5):341--358, 2011.

\bibitem{Bovolo2015}
Francesca Bovolo and Lorenzo Bruzzone.
\newblock The time variable in data fusion: A change detection perspective.
\newblock {\em IEEE Geoscience and Remote Sensing Magazine}, 3:8--26, 9 2015.

\bibitem{campos2020understanding}
Manuel Campos-Taberner, Francisco~Javier Garc{\'\i}a-Haro, Beatriz
  Mart{\'\i}nez, Emma Izquierdo-Verdiguier, Clement Atzberger, Gustau
  Camps-Valls, and Mar{\'\i}a~Amparo Gilabert.
\newblock Understanding deep learning in land use classification based on
  sentinel-2 time series.
\newblock {\em Scientific reports}, 10(1):1--12, 2020.

\bibitem{Delegido2011}
Jesús Delegido, Jochem Verrelst, Luis Alonso, and José Moreno.
\newblock Evaluation of sentinel-2 red-edge bands for empirical estimation of
  green {LAI} and chlorophyll content.
\newblock {\em Sensors}, 11:7063--7081, 7 2011.

\bibitem{Donahue2017}
Jeff Donahue, Lisa~Anne Hendricks, Marcus Rohrbach, Subhashini Venugopalan,
  Sergio Guadarrama, Kate Saenko, and Trevor Darrell.
\newblock Long-term recurrent convolutional networks for visual recognition and
  description.
\newblock {\em IEEE Transactions on Pattern Analysis and Machine Intelligence},
  39:677--691, 11 2017.

\bibitem{Ghamisi2019}
Pedram Ghamisi, Behnood Rasti, Naoto Yokoya, Qunming Wang, Bernhard Hofle,
  Lorenzo Bruzzone, Francesca Bovolo, Mingmin Chi, Katharina Anders, Richard
  Gloaguen, Peter~M. Atkinson, and Jon~Atli Benediktsson.
\newblock Multisource and multitemporal data fusion in remote sensing: A
  comprehensive review of the state of the art.
\newblock {\em IEEE Geoscience and Remote Sensing Magazine}, 7:6--39, 3 2019.

\bibitem{Ghosh2021CalCROP21AG}
Rahul Ghosh, Praveen Ravirathinam, Xiaowei Jia, Ankush Khandelwal, David~J.
  Mulla, and Vipin Kumar.
\newblock Calcrop21: A georeferenced multi-spectral dataset of satellite
  imagery and crop labels.
\newblock {\em 2021 IEEE International Conference on Big Data (Big Data)},
  pages 1625--1632, 2021.

\bibitem{GITELSON1996}
Anatoly~A. Gitelson, Yoram~J. Kaufman, and Mark~N. Merzlyak.
\newblock Use of a green channel in remote sensing of global vegetation from
  eos-modis.
\newblock {\em Remote Sensing of Environment}, 58(3):289--298, 1996.

\bibitem{GORELICK2017}
N.~Gorelick, M.~Hancher, M.~Dixon, S.~Ilyushchenko, D.~Thau, and R.~Moore.
\newblock Google earth engine: Planetary-scale geospatial analysis for
  everyone.
\newblock {\em Remote Sensing of Environment}, 2017.

\bibitem{He2016}
Kaiming He, Xiangyu Zhang, Shaoqing Ren, and Jian Sun.
\newblock Deep residual learning for image recognition.
\newblock In {\em IEEE Conference on Computer Vision and Pattern Recognition},
  pages 770--778, 12 2016.

\bibitem{Helber2019}
Patrick Helber, Benjamin Bischke, Andreas Dengel, and Damian Borth.
\newblock Eurosat: A novel dataset and deep learning benchmark for land use and
  land cover classification.
\newblock {\em IEEE Journal of Selected Topics in Applied Earth Observations
  and Remote Sensing}, 12:2217--2226, 7 2019.

\bibitem{Huang2016DenseNet}
Gao Huang, Zhuang Liu, and Kilian~Q. Weinberger.
\newblock Densely connected convolutional networks.
\newblock {\em CoRR}, abs/1608.06993, 2016.

\bibitem{Isaac2017}
Ebenezer Isaac, K.~S. Easwarakumar, and Joseph Isaac.
\newblock Urban landcover classification from multispectral image data using
  optimized adaboosted random forests.
\newblock {\em Remote Sensing Letters}, 8:350--359, 4 2017.

\bibitem{JIA2013}
Xiuping Jia, Bor-Chen Kuo, and Melba~M. Crawford.
\newblock Feature mining for hyperspectral image classification.
\newblock {\em Proceedings of the IEEE}, 101(3):676--697, March 2013.

\bibitem{Jung2021}
Jinha Jung, Murilo Maeda, Anjin Chang, Mahendra Bhandari, Akash Ashapure, and
  Juan Landivar-Bowles.
\newblock The potential of remote sensing and artificial intelligence as tools
  to improve the resilience of agriculture production systems.
\newblock {\em Current Opinion in Biotechnology}, 70:15--22, 8 2021.

\bibitem{KAMILARIS2018}
A.~Kamilaris and F.X. Prenafeta-Boldú.
\newblock Deep learning in agriculture: A survey.
\newblock {\em Computers and Electronics in Agriculture}, 147:70--90, 2018.

\bibitem{dKerselaersLIPS2019}
Eva Kerselaers and Gregor Levin.
\newblock Applying lpis data to assess loss of agricultural land –
  experiences from flanders and denmark.
\newblock {\em Geografisk Tidsskrift-Danish Journal of Geography},
  119(1):17--29, 2019.

\bibitem{Khatami2016}
Reza Khatami, Giorgos Mountrakis, and Stephen~V. Stehman.
\newblock A meta-analysis of remote sensing research on supervised pixel-based
  land-cover image classification processes: General guidelines for
  practitioners and future research.
\newblock {\em Remote Sensing of Environment}, 177:89--100, 5 2016.

\bibitem{kondmann2021denethor}
Lukas Kondmann, Aysim Toker, Marc Ru{\ss}wurm, Andr{\'e}s Camero, Devis
  Peressuti, Grega Milcinski, Pierre-Philippe Mathieu, Nicolas Long{\'e}p{\'e},
  Timothy Davis, Giovanni Marchisio, Laura Leal-Taix{\'e}, and Xiao~Xiang Zhu.
\newblock {DENETHOR}: The dynamicearth{NET} dataset for harmonized,
  inter-operable, analysis-ready, daily crop monitoring from space.
\newblock In {\em Thirty-fifth Conference on Neural Information Processing
  Systems Datasets and Benchmarks Track (Round 2)}, 2021.

\bibitem{Kumar2018}
Lalit Kumar and Onisimo Mutanga.
\newblock Google earth engine applications since inception: Usage, trends, and
  potential.
\newblock {\em Remote Sensing}, 10:1509, 10 2018.

\bibitem{Kussul2017}
Nataliia Kussul, Mykola Lavreniuk, Sergii Skakun, and Andrii Shelestov.
\newblock Deep learning classification of land cover and crop types using
  remote sensing data.
\newblock {\em IEEE Geoscience and Remote Sensing Letters}, 14:778--782, 5
  2017.

\bibitem{li2021crop}
Minghui Li, Junjie Guo, Tao Ren, Gongwen Luo, Qirong Shen, Jianwei Lu, Shiwei
  Guo, and Ning Ling.
\newblock Crop rotation history constrains soil biodiversity and
  multifunctionality relationships.
\newblock {\em Agriculture, Ecosystems \& Environment}, 319:107550, 2021.

\bibitem{Li2018}
Ruirui Li, Wenjie Liu, Lei Yang, Shihao Sun, Wei Hu, Fan Zhang, and Wei Li.
\newblock {DeepUNet}: A deep fully convolutional network for pixel-level
  sea-land segmentation.
\newblock {\em IEEE Journal of Selected Topics in Applied Earth Observations
  and Remote Sensing}, 11:3954--3962, 11 2018.

\bibitem{Masoud2020}
Khairiya~Mudrik Masoud, Claudio Persello, and Valentyn~A. Tolpekin.
\newblock Delineation of agricultural field boundaries from sentinel-2 images
  using a novel super-resolution contour detector based on fully convolutional
  networks.
\newblock {\em Remote Sensing}, 12:59, 1 2020.

\bibitem{MAZZIA2020}
Vittorio Mazzia, Aleem Khaliq, and Marcello Chiaberge.
\newblock Improvement in land cover and crop classification based on temporal
  features learning from sentinel-2 data using recurrent-convolutional neural
  network {(R-CNN)}.
\newblock {\em Applied Sciences}, 10:238, 1 2020.

\bibitem{Merzlyak1999}
Mark~N Merzlyak, Anatoly~A Gitelson, Olga~B Chivkunova, and Victor~Yu Rakitin.
\newblock Non-destructive optical detection of pigment changes during leaf
  senescence and fruit ripening.
\newblock {\em Physiologia Plantarum}, 106:135--141, 1999.

\bibitem{NRSC}
{Natural Resources Canada}.
\newblock Land cover and land use.
\newblock
  \url{https://www.nrcan.gc.ca/maps-tools-and-publications/satellite-imagery-and-air-photos/remote-sensing-tutorials/land-cover-land-use/},
  2015.

\bibitem{OKSUZ2019136}
Ilkay Oksuz, Bram Ruijsink, Esther Puyol-Antón, James~R. Clough, Gastao Cruz,
  Aurelien Bustin, Claudia Prieto, Rene Botnar, Daniel Rueckert, Julia~A.
  Schnabel, and Andrew~P. King.
\newblock Automatic cnn-based detection of cardiac mr motion artefacts using
  k-space data augmentation and curriculum learning.
\newblock {\em Medical Image Analysis}, 55:136--147, 2019.

\bibitem{omalley2019kerastuner}
Tom O'Malley, Elie Bursztein, James Long, Fran\c{c}ois Chollet, Haifeng Jin,
  Luca Invernizzi, et~al.
\newblock Keras {Tuner}.
\newblock \url{https://github.com/keras-team/keras-tuner}, 2019.

\bibitem{RONDEAUX199695}
Geneviève Rondeaux, Michael Steven, and Frédéric Baret.
\newblock Optimization of soil-adjusted vegetation indices.
\newblock {\em Remote Sensing of Environment}, 55(2):95--107, 1996.

\bibitem{ROUSE1974}
{J.W. Jr.} {Rouse}, R.H. {Haas}, J.A. {Schell}, and D.W. {Deering}.
\newblock {\em {Monitoring Vegetation Systems in the {Great Plains} with
  {Erts}}}, volume 351, page 309.
\newblock NASA Special Publication, 1974.

\bibitem{SHI2015}
Xingjian Shi, Zhourong Chen, Hao Wang, Dit-Yan Yeung, Wai-kin Wong, and
  Wang-chun Woo.
\newblock Convolutional lstm network: A machine learning approach for
  precipitation nowcasting.
\newblock In {\em Proceedings of the 28th International Conference on Neural
  Information Processing Systems - Volume 1}, NIPS'15, page 802–810,
  Cambridge, MA, USA, 2015. MIT Press.

\bibitem{Song2019}
Hunsoo Song, Yonghyun Kim, and Yongil Kim.
\newblock A patch-based light convolutional neural network for land-cover
  mapping using landsat-8 images.
\newblock {\em Remote Sensing}, 11:114, 1 2019.

\bibitem{Sumbul2021}
Gencer Sumbul, Arne de~Wall, Tristan Kreuziger, Filipe Marcelino, Hugo Costa,
  Pedro Benevides, Mario Caetano, Beg{\"{u}}m Demir, and Volker Markl.
\newblock Bigearthnet-mm: {A} large scale multi-modal multi-label benchmark
  archive for remote sensing image classification and retrieval.
\newblock {\em CoRR}, abs/2105.07921, 2021.

\bibitem{Tan2019EfficientNet}
Mingxing Tan and Quoc~V. Le.
\newblock Efficientnet: Rethinking model scaling for convolutional neural
  networks.
\newblock {\em CoRR}, abs/1905.11946, 2019.

\bibitem{TEIMOURI2019}
Nima Teimouri, Mads Dyrmann, and Rasmus~Nyholm Jørgensen.
\newblock A {Novel} {Spatio}-{Temporal} {FCN}-{LSTM} {Network} for
  {Recognizing} {Various} {Crop} {Types} {Using} {Multi}-{Temporal} {Radar}
  {Images}.
\newblock {\em Remote Sensing}, 11(8):990, April 2019.

\bibitem{SENTINEL2}
{The European Space Agency}.
\newblock Radiometric resolutions.
\newblock
  \url{https://sentinel.esa.int/web/sentinel/user-guides/sentinel-2-msi/resolutions/radiometric},
  2021.

\bibitem{tseng2021cropharvest}
Gabriel Tseng, Ivan Zvonkov, Catherine~Lilian Nakalembe, and Hannah Kerner.
\newblock Cropharvest: A global dataset for crop-type classification.
\newblock In {\em Thirty-fifth Conference on Neural Information Processing
  Systems Datasets and Benchmarks Track (Round 2)}, 2021.

\bibitem{WANG2018}
Qunming Wang and Peter~M. Atkinson.
\newblock Spatio-temporal fusion for daily sentinel-2 images.
\newblock {\em Remote Sensing of Environment}, 204:31--42, 2018.

\bibitem{Wessel2018}
Mathias Wessel, Melanie Brandmeier, and Dirk Tiede.
\newblock Evaluation of different machine learning algorithms for scalable
  classification of tree types and tree species based on sentinel-2 data.
\newblock {\em Remote Sensing}, 10:1419, 9 2018.

\bibitem{Xue2017}
Jinru Xue and Baofeng Su.
\newblock Significant remote sensing vegetation indices: A review of
  developments and applications.
\newblock {\em Journal of Sensors}, 2017, 2017.

\bibitem{Yang2018}
Chenghai Yang.
\newblock High resolution satellite imaging sensors for precision agriculture.
\newblock {\em Frontiers of Agricultural Science and Engineering}, 5:393--405,
  11 2018.

\bibitem{Zhang2016}
Liangpei Zhang, Lefei Zhang, and Bo~Du.
\newblock Deep learning for remote sensing data: A technical tutorial on the
  state of the art.
\newblock {\em IEEE Geoscience and Remote Sensing Magazine}, 4:22--40, 6 2016.

\bibitem{zhao2021evaluation}
Hongwei Zhao, Sibo Duan, Jia Liu, Liang Sun, and Louis Reymondin.
\newblock Evaluation of five deep learning models for crop type mapping using
  sentinel-2 time series images with missing information.
\newblock {\em Remote Sensing}, 13(14):2790, 2021.

\bibitem{Zhu2017}
Xiao~Xiang Zhu, Devis Tuia, Lichao Mou, Gui~Song Xia, Liangpei Zhang, Feng Xu,
  and Friedrich Fraundorfer.
\newblock Deep learning in remote sensing: A comprehensive review and list of
  resources.
\newblock {\em IEEE Geoscience and Remote Sensing Magazine}, 5:8--36, 12 2017.

\bibitem{Zunair20203dcnn}
Hasib Zunair, Aimon Rahman, Nabeel Mohammed, and Joseph~Paul Cohen.
\newblock Uniformizing techniques to process ct scans with 3d cnns for
  tuberculosis prediction.
\newblock In Islem Rekik, Ehsan Adeli, Sang~Hyun Park, and Maria del~C.
  Vald{\'e}s~Hern{\'a}ndez, editors, {\em Predictive Intelligence in Medicine},
  pages 156--168. Springer International Publishing, 2020.

\end{thebibliography}

\newpage
\appendix

\section{Appendix}

\subsection{Sentinel-2 reflectance bands}\label{sentinel_bands}

\begin{table}[H]
    \caption{List of the main spectral bands captured by Sentinel-2 and their respective wavelengths.}\label{tab:sentinel2_bands}
    \centering 
        \begin{tabular}{c c c c}
        \hline
            Band name &   Spatial resolution (pixel/m)  &   Central wavelength (nm) &   Description \\
        \hline
            B1      &   60              &       444         &   Aerosols    \\
            B2      &   10              &       497         &   Blue        \\
            B3      &   10              &       560         &   Green       \\
            B4      &   10              &       665         &   Red         \\
            B5      &   20              &       704         &   Red Edge 1  \\
            B6      &   20              &       740         &   Red Edge 2  \\
            B7      &   20              &       783         &   Red Edge 3  \\
            B8      &   10              &       835         &   NIR         \\
            B8A     &   20              &       865         &   Red Edge 4  \\
            B9      &   60              &       945         &   Water vapor \\
            B11     &   20              &       1614        &   SWIR1       \\
            B12     &   20              &       2202        &   SWIR2       \\
        \hline
        \end{tabular}
\end{table}

\subsection{\textit{Canadian Annual Crop Inventory} accuracy table}\label{ACI_accuracy_table}

\begin{table}[H]
    \caption{Summary of the ACI crop class accuracy for each year and province.}\label{tab:ACI_accuracy_tab}
    \centering 
        \begin{tabular}{l c c c c c c}
        \hline
           Province                 &   2016    &   2017    &   2018    &   2019    &   2020    &   Average (2016-2020) \\
        \hline
            Alberta                 &   90.83   &   94.15   &   91.95   &   91.29   &   88.94   &       91.43   \\
            British Columbia        &   82.27   &   92.79   &   93.09   &   89.35   &   85.16   &       88.53   \\
            Manitoba                &   92.44   &   93.10   &   94.61   &   94.27   &   93.47   &       93.58   \\
            New Brunswick           &   89.66   &   84.29   &   88.83   &   91.90   &   95.74   &       90.08   \\
            Newfoundland            &   94.51   &   91.83   &   93.84   &   91.00   &   95.08   &       93.25   \\
            Nova Scotia             &   90.59   &   89.49   &   92.50   &   89.10   &   NaN     &       90.42   \\
            Ontario                 &   88.98   &   85.36   &   91.99   &   85.64   &   88.26   &       88.05   \\
            Prince Edward Island    &   82.44   &   91.61   &   81.92   &   89.78   &   85.85   &       86.32   \\
            Quebec                  &   91.17   &   90.26   &   92.28   &   91.80   &   91.2    &       91.34   \\
            Saskatchewan            &   92.26   &   93.71   &   91.65   &   91.63   &   93.87   &       92.62   \\
        \hline
            Average (all provinces) &   89.52   &   90.66   &   91.27   &   90.58   &   90.8    &  \textbf{90.56} \\
      
        \hline
        
        \end{tabular}
        
          {\textit{Source: Annual Crop Inventory - Data Product Specification - As of June 7th, 2022. Missing values are imputed using the mean of each row.} \cite{ACI} \par}
        
\end{table}

\subsection{Definition of vegetation indices}\label{definition_vegetation_indices}

\paragraph{NDVI}
The \textit{Normalized Difference Vegetation Index} (NDVI) is one of the most predominantly used indicators of plant growth and health. It correlates with the amount of chlorophyll emitted by a plant \cite{ROUSE1974}. NDVI is defined as:  

\begin{equation}
    \label{eq:NDVI} 
        NDVI = \frac{ \rho_{NIR}-\rho_{red}}{\rho_{NIR}+\rho_{red}}
\end{equation}

Where $\rho_{NIR}$ is equal to the NIR band and $\rho_{red}$ is the visible red band (represented by bands B8 and B4 in Table \ref{tab:sentinel2_bands}, respectively).

\paragraph{NDVI45}
The \textit{NDVI45} vegetation index is a revised version of the NDVI developed by \cite{Delegido2011}. It is strongly correlated with the leaf area index (LAI), which is an estimate of the amount of biomass and vegetative evapotranspiration, and provides information regarding the structure of a canopy. NDVI45 is defined as:  

\begin{equation}
    \label{eq:NDVI45} 
        NDVI45 = \frac{ R_{704}-\rho_{red}}{R_{704}+\rho_{red}}
\end{equation}

Where $R_{704}$ is the NIR spectral band centered at 704 nm (B5). 

\paragraph{GNDVI} The \textit{Green Normalized Difference Vegetation Index} (GNDVI) was shown to  correlate to the rate of photosynthesis and is used to monitor plant stress \cite{GITELSON1996}. GNDVI is calculated in a way that is analogous to NDVI, however the red band is replaced by the green band ($\rho_{green}$):  

\begin{equation}
    \label{eq:GNDVI} 
        GNDVI = \frac{\rho_{NIR}-\rho_{green}}{\rho_{NIR}+\rho_{green}} 
\end{equation}

\paragraph{PSRI} The \textit{Plant Senescence Reflectance Index} (PSRI) is used to measure the onset, the stage, and the relative rates of the senescence or ripening of a crop cover. An increase in PSRI indicates heightened canopy stress \cite{Merzlyak1999}. PSRI is defined by the equation:  

\begin{equation}
    \label{eq:PSRI} 
        PSRI = \frac{\rho_{red}-\rho_{blue}}{R_{750}} 
\end{equation}

Where $R_{750}$ is the NIR spectral band centered at 750 nm (B6).

\paragraph{OSAVI} In conditions when vegetation is low and soil properties are unknown, indices like the NDVI can be subject to bias due to high levels of reflection. The \textit{Optimized Soil-Adjusted Variation Index} (OSAVI) was created to provide an estimate of biomass that is more resilient when faced with soil and atmospheric effects \cite{RONDEAUX199695}.

\begin{equation}
    \label{eq:OSAVI} 
         OSAVI = \frac{\rho_{NIR}-\rho_{red}}{\rho_{NIR} + \rho{red} + 0.16} 
\end{equation}

\subsection{An example image from the \textit{Canadian Cropland Dataset}.}\label{barley_example}

\begin{figure}[h!]
    \centering
    \includegraphics[width=1.0\textwidth]{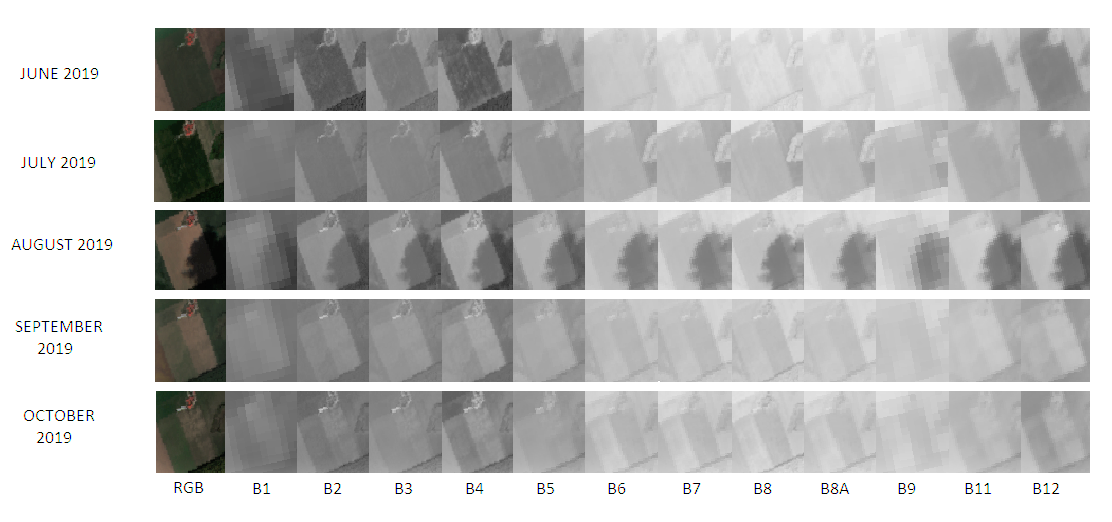}
    \caption{Example of a Sentinel-2 image for a barley field in Ontario, Canada. The different bands (B1 to B12) represent the median of the pixels for each month.} \label{fig:barley_example}
\end{figure}

\subsection{Definition of performance metrics}\label{performance_metrics}

\paragraph{Accuracy} To evaluate the performance of each classification algorithm, a selection of performance metrics were used. \textit{Accuracy} (equation \ref{eq:accuracy_def}) is an important metric that represents the capacity of an algorithm to correctly classify instances. Although it is a powerful indicator of overall performance, accuracy alone is not enough to determine the strength of an algorithm and whether it has correctly learned the task at hand. Accuracy is defined as: 

\begin{equation}
\label{eq:accuracy_def}
Accuracy = \frac{\Sigma TP+ \Sigma TN}{\Sigma TP+ \Sigma TN +\Sigma FP+ \Sigma FN}, 
\end{equation}

where $TP$ and $TN$ are the number of true positive and true negative classifications, respectively; and $FP$ and $FN$ are the number of false positive and false negative classifications, respectively. Accuracy, like all metrics, is often multiplied by 100 to yield a percentage.  

\paragraph{Precision} \textit{Precision} (equation \ref{eq:precision_def}), also referred to as positive predictive value (PPV), is used to determine the capacity of an algorithm to correctly identify positive cases with respect to all the cases the algorithm has classified as positive. It is calculated by dividing the number of true positives by the number of predicted positives, which itself is a sum of TP and FP. Precision is defined as: 

\begin{equation}
\label{eq:precision_def}
Precision = \frac{\Sigma TP}{\Sigma TP+\Sigma FP}. 
\end{equation}

Precision is an indicator of how reproducible and repeatable a measurement is under unchanged conditions and is used to evaluate the exactness of a model.

\paragraph{Recall} \textit{Recall} (equation \ref{eq:recall_def}) is the fraction of relevant instances that have been correctly identified (TP) over the total amount of relevant instances (TP and FN). Recall and precision are typically used in unison to report the performance of a classification system. Precision indicates the quality of the positive prediction capability of the model, while recall indicates the completeness or quantity of correct predictions with respect to all positive instances present. High precision would mean that the algorithm returned a greater amount of relevant results than irrelevant ones, while a high recall value would mean that the algorithm returned most of the relevant results.

\begin{equation}
Recall = \frac{\Sigma TP}{\Sigma TP+\Sigma FN} \label{eq:recall_def}
\end{equation}

\paragraph{F1-Score} \textit{F1-Score} or \textit{F-measure}  (equation \ref{eq:fscore_def}) 
is a metric that combines both precision and recall into a single encompassing metric. This weighted average is bounded between 0 - representing the worst classification and 1 - representing the best. 

\begin{equation}
F1-Score = 2 \times \frac{Precision \times Recall}{Precision + Recall} \label{eq:fscore_def}
\end{equation}

\subsection{LRCN performance metrics (RGB 2019 dataset)}\label{LRCN_perf_metrics}

\begin{table}[ht]
    \caption{LRCN performance metrics (RGB 2019 dataset)} \label{tab: LRCN_perf_metrics} 
        \centering
        \begin{tabular}{l c c c c}
            \toprule
                                  &      \multicolumn{4}{c}{Number of hidden LSTM units in the last layer.}        \\
            \cmidrule(r){2-5}
                 Metrics           &     32        &   64      &   128     &   256     \\
            \hline
                 Avg. Precision    &     0.59      &   0.61    &   0.59    &   0.62       \\
                 Avg. Recall       &     0.62      &   0.64    &   0.62    &   0.62       \\
                 Avg. F1-Score     &     0.60      &   0.62    &   0.60    &   0.61       \\
                 Avg. Accuracy     &     0.76      &   0.77    &   0.74    &   0.74       \\ 
            \hline
        \end{tabular}
\end{table}

\newpage

\subsection{Complete summary of performance metrics for all models }\label{Model_perf_metrics}

\begin{table}[ht]
\caption{Model performance metrics (RGB 2019 dataset)*}
\label{tab:Model_perf_metrics} 
\centering
\begin{tabular}{@{}lllll@{}}
\toprule
\textbf{Models/Metrics}        & \textbf{Precision} & \textbf{Recall} & \textbf{F1-Score} & \textbf{Accuracy}      \\ \hline
\textit{\textbf{Dynamic}}      &                    &                 &                   &                        \\
\textbf{LRCN-64 (RGB)}         & 0.610 ± 0.017      & 0.637 ± 0.015   & 0.617 ± 0.012     & \textbf{0.774 ± 0.014} \\
\textbf{LRCN-64 (GNDVI)}       & 0.030 ± 0.000      & 0.100 ± 0.000   & 0.040 ± 0.000     & 0.277 ± 0.006          \\
\textbf{LRCN-64 (NDVI)}        & 0.030 ± 0.000      & 0.100 ± 0.000   & 0.040 ± 0.000     & 0.270 ± 0.000          \\
\textbf{LRCN-64 (NDVI45)}      & 0.123 ± 0.162      & 0.180 ± 0.139   & 0.127 ± 0.150     & 0.313 ± 0.101          \\
\textbf{LRCN-64 (OSAVI)}       & 0.140 ± 0.191      & 0.163 ± 0.110   & 0.117 ± 0.133     & 0.300 ± 0.087          \\
\textbf{LRCN-64 (PSRI)}        & 0.450 ± 0.010      & 0.467 ± 0.012   & 0.453 ± 0.012     & 0.563 ± 0.012          \\
\textbf{3D-CNN (RGB)}          & 0.610 ± 0.026      & 0.620 ± 0.000   & 0.607 ± 0.012     & \textbf{0.773 ± 0.012} \\
\textbf{3D-CNN (GNDVI)}        & 0.287 ± 0.085      & 0.250 ± 0.066   & 0.200 ± 0.069     & 0.313 ± 0.067          \\
\textbf{3D-CNN (NDVI)}         & 0.360 ± 0.113      & 0.373 ± 0.093   & 0.347 ± 0.104     & 0.470 ± 0.100          \\
\textbf{3D-CNN (NDVI45)}       & 0.467 ± 0.065      & 0.387 ± 0.031   & 0.377 ± 0.040     & 0.530 ± 0.026          \\
\textbf{3D-CNN (OSAVI)}        & 0.433 ± 0.151      & 0.447 ± 0.136   & 0.417 ± 0.153     & 0.570 ± 0.114          \\
\textbf{3D-CNN (PSRI)}         & 0.447 ± 0.040      & 0.350 ± 0.087   & 0.347 ± 0.087     & 0.477 ± 0.078          \\
\textit{\textbf{Static}}       &                    &                 &                   &                        \\
\textbf{DENSENET-121 (RGB)}    & 0.327 ± 0.064      & 0.330 ± 0.030   & 0.273 ± 0.058     & \textbf{0.403 ± 0.045} \\
\textbf{DENSENET-121 (GNDVI)}  & 0.057 ± 0.029      & 0.133 ± 0.042   & 0.060 ± 0.044     & 0.150 ± 0.053          \\
\textbf{DENSENET-121 (NDVI)}   & 0.217 ± 0.067      & 0.183 ± 0.015   & 0.140 ± 0.030     & 0.260 ± 0.026          \\
\textbf{DENSENET-121 (NDVI45)} & 0.187 ± 0.055      & 0.257 ± 0.035   & 0.180 ± 0.036     & 0.293 ± 0.032          \\
\textbf{DENSENET-121 (OSAVI)}  & 0.213 ± 0.021      & 0.210 ± 0.044   & 0.153 ± 0.038     & 0.257 ± 0.051          \\
\textbf{DENSENET-121 (PSRI)}   & 0.160 ± 0.020      & 0.197 ± 0.015   & 0.147 ± 0.012     & 0.260 ± 0.017          \\
\textbf{EFFN-B0 (RGB)}         & 0.433 ± 0.042      & 0.430 ± 0.010   & 0.410 ± 0.017     & \textbf{0.527 ± 0.012}          \\
\textbf{EFFN-B0 (GNDVI)}       & 0.427 ± 0.015      & 0.430 ± 0.010   & 0.413 ± 0.015     & 0.527 ± 0.006          \\
\textbf{EFFN-B0 (NDVI)}        & 0.343 ± 0.055      & 0.310 ± 0.052   & 0.303 ± 0.046     & 0.400 ± 0.053          \\
\textbf{EFFN-B0 (NDVI45)}      & 0.470 ± 0.017      & 0.460 ± 0.010   & 0.443 ± 0.012     & 0.543 ± 0.006          \\
\textbf{EFFN-B0 (OSAVI)}       & 0.493 ± 0.021      & 0.477 ± 0.006   & 0.463 ± 0.012     & 0.567 ± 0.006          \\
\textbf{EFFN-B0 (PSRI)}        & 0.417 ± 0.029      & 0.393 ± 0.006   & 0.380 ± 0.010     & 0.483 ± 0.006          \\
\textbf{RES-SCR (RGB)}         & 0.487 ± 0.021      & 0.473 ± 0.006   & 0.467 ± 0.012     & \textbf{0.577 ± 0.015} \\
\textbf{RES-SCR (PSRI)}        & 0.333 ± 0.042      & 0.297 ± 0.029   & 0.277 ± 0.029     & 0.393 ± 0.006          \\
\textbf{RES-SCR (NDVI)}        & 0.280 ± 0.036      & 0.293 ± 0.012   & 0.257 ± 0.021     & 0.370 ± 0.026          \\
\textbf{RES (RGB)}             & 0.587 ± 0.040      & 0.550 ± 0.044   & 0.553 ± 0.047     & \textbf{0.667 ± 0.032} \\
\textbf{RES (GNDVI)}           & 0.523 ± 0.067      & 0.423 ± 0.031   & 0.413 ± 0.040     & 0.523 ± 0.031          \\
\textbf{RES (NDVI)}            & 0.463 ± 0.031      & 0.410 ± 0.036   & 0.413 ± 0.025     & 0.517 ± 0.025          \\
\textbf{RES (NDVI45)}          & 0.480 ± 0.020      & 0.453 ± 0.015   & 0.443 ± 0.023     & 0.547 ± 0.021          \\
\textbf{RES (OSAVI)}           & 0.533 ± 0.060      & 0.530 ± 0.026   & 0.513 ± 0.040     & 0.617 ± 0.035          \\
\textbf{RES (PSRI)}            & 0.437 ± 0.046      & 0.397 ± 0.012   & 0.400 ± 0.017     & 0.483 ± 0.012          \\ \hline
\end{tabular}
 
 {\raggedright *LRCN-64 (LRCN with 64 hidden units in the last LSTM layer - see Table \ref{tab: LRCN_perf_metrics}), RES (pre-trained ResNet-50), RES-SCR (ResNet trained from scratch, EFFN-B0 (EfficientNet-B0) - See section \textit{Static and Dynamic image classification}). The presented scores are average ± standard deviation (SD) of triplicate experiments with different starting random seeds. Results in \textbf{bold} are the average accuracy of the models trained with the RGB 2019 dataset. All models were implemented using Keras and Tensorflow. \par}
 
\end{table}

\newpage

\subsection{Canadian  Cropland Dataset - License}\label{dataset_license}

\textbf{Montreal Data License (MDL)}

The following licensing language is made available under CC-BY4.  Attribution should be made to Montreal Data License (MDL), or License language based on Montreal Data License. 

The authors are not legal advisors to the individuals and entities making use of these licensing terms. The licensing terms can be combined as needed to match the rights conferred by the licensor. 

The language below assumes that all rights are granted, however each right should be conferred or not based on the users intent. 

\textit{Data License for use in AI and ML: }

This license covers the Data made available by Licensor to you (License) under the following terms. Licensees use of the data consists acceptance of the terms of this license agreement (License). 

\begin{enumerate}

\item \textit{\textbf{Definitions}}
    \begin{enumerate}
        \item \textit{Data} means the informational content (individually or as a whole) made available by Licensor. 
        
        \item \textit{Model} means machine-learning or artificial-intelligence based algorithms, or assemblies thereof that,  in combination with different techniques,  may be used to obtain certain results.  Without limitation, such results can be insights on past data patterns, predictions on future trends or more abstract results. 
        
        \item \textit{Output} means the results of operating a Trained Model as embodied in informational content resulting therefrom. 
        
        \item \textit{Representation} is  a  transformation  of  a  piece  of  data  into  a  different  form.   Good representations can be used as input to perform useful tasks. 
        
        \item \textit{Labelled Data} means the associated metadata and informational content derived from Data which identify, comment or otherwise derive information from Data, such as tags and labels. 
        
        \item \textit{Licensor} means the individual or entity making the Data available to you. 
        
        \item \textit{Third Parties} means individuals or entities that are not under common control with Licensee. 
        
        \item \textit{Train} means to expose an Untrained Model to the Data in order to adjust the weights, hyperparameters and/or structure thereof. 
        
        \item \textit{Trained Model} means a Model that is exposed to Data such that its weights, parameters and architecture embody insights from the Data. 
        
        \item \textit{Untrained  Model} means  Model  that  is  conceived  and  reduced  to  practice  as  to  its structure, components and architecture but that has not been trained on Data such that its weights, parameters and architecture do not embody insights from the Data. 
    \end{enumerate}

\item \textit{\textbf{General Clauses} }
    \begin{enumerate}

        \item Unless otherwise agreed in writing by the parties,  the data is licensed as is and as available.  Licensor excludes all representations, warranties, obligations, and liabilities, whether express or implied, to the maximum extent permitted by law. 

        \item Nothing in this License permits Licensee to make use of Licensors trademarks, tradenames,  logos or to otherwise suggest endorsement or misrepresent the relationship between the parties. 

        \item The  rights  granted  under  this  license  are  deemed  to  be  non-exclusive,  worldwide, perpetual and irrevocable, unless otherwise specified in writing by Licensor. 

        \item Without limiting Licensees rights available under applicable law,  all rights not expressly granted hereunder are hereby reserved by Licensor. The Data and the database under which it is made available remain the property of Licensor (and/or its affiliates or licensors). 

        \item This  license  shall  be  terminated  upon  any  breach  by  Licensee  of  the  terms  of  this License. 
    \end{enumerate}

\item \textit{\textbf{Licensed Rights to the Data} }
    \begin{enumerate}
        \item Licensor hereby grants the following rights to Licensee with respect to making use of the Data itself. 
            \begin{enumerate}
                \item Access the Data, where access means to access, view and/or download the Data to  view  it  and  evaluate  it  (evaluation  algorithms  may  be  exposed  to  it,  but  no Untrained Models). 

                \item Creation of Tagged Data. 

                \item Distribute the Data, i.e.  to make all or part of the Data available to Third Parties under the same terms as those of this License. 

                \item Creation of a Representation of the Data. 
            \end{enumerate}
    \end{enumerate}

\item \textit{\textbf{Licensed Rights in Conjunction with Models}} 
    \begin{enumerate}
        \item Licensor hereby \textbf{grants} the following rights to Licensee with respect to making use of the Data in conjunction with Models. 
        \begin{enumerate}
            \item \textit{Benchmark: } To access the Data, use the Data as training data to evaluate the efficiency of different Untrained Models,  algorithms and structures,  but excludes reuse of the Trained Model, except to show the results of the Training.  This includes the right to use the dataset to measure performance of a Trained or Un-trained Model, without however having the right to carry-over weights, code or architecture or implement any modifications resulting from the Evaluation. 

            \item \textit{Research: }To access the Data, use the Data to create or improve Models, but with-out the right to use the Output or resulting Trained Model for any purpose other than evaluating the Model Research under the same terms. 

            \item \textit{Publish:  }To make available to Third Parties the Models resulting from Research, provided however that third parties accessing such Trained Models have the right to use them for Research or Publication only. 

            \item \textit{Internal Use:  }To access the Data, use the Data to create or improve Models and resulting  Output,  but  without  the  right  to  Output  Commercialization  or  Model Commercialization.  The Output can be used internally for any purpose, but not made available to Third Parties or for their benefit. 
            
            \end{enumerate}

        \item The rights granted in (a) above \textbf{exclude} the following rights with respect to making use of the Data in conjunction with Models: 
            \begin {enumerate}
            \item \textit{Output Commercialization: }To access the Data, use the Data to create or improve Models and resulting Output, with the right to make the Output available to Third Parties or to use it for their benefit, without the right to Model Commercialization. 
            
            \item \textit{Model Commercialization:} Make a Trained Model itself available to a Third Party, or embodying the Trained Model in a product or service, with or without direct access to the Output for such Third Party. 
            \end{enumerate}
        
        \end{enumerate}

\item \textit{\textbf{Attribution and Notice Attribution and Notice}} \\ 
The origin of the Data and notices included with the Data shall be made available to Third Parties to whom the Data, Output and/Model have been made available. Any distribution of all or part of the Data shall be done under the same terms as those of this License. Licensee shall make commercially reasonable efforts to link to the source of the Data.  If so indicated by the Licensor in writing alongside the Data that the use shall be deemed confidential, then Licensee shall not publicly refer to Licensor and/or the source of the Data. 

\end{enumerate}

\newpage 
\subsection{Canadian Cropland Dataset - Datasheet}\label{canadian_crop_datasheet}

\begin{multicols}{2}
\setlength{\columnsep}{1cm}
\justifying

\subsubsection{Motivation}

\textit{\textbf{For what purpose was the dataset created? Was there a specific task in mind? Was there a specific gap that needed to be filled? Please provide a description.}}

This dataset was created to provide images of different types of Canadian croplands to study cropland classification using geo-referenced and multitemporal data. This dataset was created to help improve the classification of Canadian croplands using remote sensing imagery under multiple possible scenarios (in a temporal and/or static image classification context).  

\textit{\textbf{Who created the dataset (e.g., which team, research group) and on behalf of which entity (e.g., company, institution, organization)?}}

The dataset was created in 2021 by a research scientist (Dr. Etienne Lord), a doctoral student (Amanda A. Boatswain Jacques) of Agriculture and Agri-Food Canada (AAFC), and a professor at the University of Quebec in Montreal (UQAM) (Dr. Abdoulaye Baniré Diallo). 

\textit{\textbf{Who funded the creation of the dataset? If there is an associated grant, please provide the name of the grant or\textbackslash and the grant number.}}

The construction of this dataset was funded by \href{https://agriculture.canada.ca/en/about-our-department}{AAFC}, the department of the Government of Canada that is responsible for policies governing agricultural production, processing and marketing of all farm, food and agri-based products. This project is associated with the Smart Land Management Approach Grant.   This project was also funded by a UQAM internal grant. 

\textit{\textbf{Any other comments?}} N/A.

\subsubsection{Composition}

\textit{\textbf{What do the instances that comprise the dataset represent (e.g., documents, photos, people, countries)? Are there multiple types of instances (e.g., movies, users, and ratings; people and interactions between them; nodes and edges)? Please provide a description.}}

The instances that comprise this dataset are satellite images of Canadian croplands that are sorted according to crop type, location throughout the country and date.

\textit{\textbf{How many instances are there in total (of each type, if appropriate)?}}

The cleaned dataset contains a total of 78,536 distinct instances from 10 predominant crop types. The number of instances per class is reported below: 


\begin{center}
\begin{tabular}{|l|c|}
    \hline 
        \textbf{Crop types}      &	\textbf{Instances}  \\
    \hline 
        CORN                    &       12878           \\
        CANOLA                  &       11366	        \\
        PASTURE	                &       8797            \\
        SPRING WHEAT	        &       8229            \\
        ORCHARD                 &       6594            \\
        BARLEY	                &       5382            \\
        POTATO                  &	    5294            \\
        MIXEDWOOD               &	    4981            \\
        OAT	                    &       4807            \\
     \hline 
        \textbf{TOTAL}          &   \textbf{78536}      \\
    \hline

     \hline 
\end{tabular}
\end{center}

The raw version of the dataset may have up to 46 classes per year. 

\textit{\textbf{Does the dataset contain all possible instances or is it a sample (not necessarily random) of instances from a larger set? If the dataset is a sample, then what is the larger set? Is the sample representative of the larger set (e.g., geographic coverage)? If so, please describe how this representativeness was validated  and verified. If it is not representative of the larger set, please describe why not (e.g., to cover a more diverse range of instances, because instances were withheld or unavailable).}}

The dataset is a sample of the different crop types within different months and year. However, the current dataset is captured in rolling manner. 

\textit{\textbf{What data does each instance consist of? “Raw” data (e.g., unprocessed text or images) or features? In either case, please provide a description.}}

Each instance of this dataset consists of a \href{https://sentinel.esa.int/web/sentinel/missions/sentinel-2}{Sentinel-2} satellite image for a single geographical point taken at a specific time period. These images were scraped using the open-source geospatial analysis tool \href{https://earthengine.google.com/}{Google Earth Engine} (GEE). Each image is composed of 12 spectral bands which range from the Visible (VNIR) and Near Infra-Red (NIR) to the Short Wave Infra-Red (SWIR) wavelengths. Each point corresponds to the center of a Canadian agricultural field of a specific crop type. The images represent an area of 640 x 640 m, captured at a resolution of 10 m/pixel. 

\textit{\textbf{Is there a label or target associated with each instance? If so, please provide a description.}}

Each instance is associated with a label that dictates the type of crop present within the image. This label is present in the filename of the instance. For example, \textit{\texttt{POINT\_2\_201909\_AB\_BARLEY}} corresponds to the geographical location with a point ID of 2, was retrieved in September 2019 from the province of Alberta (AB), and belongs to the crop class BARLEY. 

\textit{\textbf{Is any information missing from individual instances? If so, please
provide a description, explaining why this information is missing (e.g.,
because it was unavailable).}}

To the best of our knowledge, we think that the information for individual instances is completed. 

\textit{\textbf{Are relationships between individual instances made explicit (e.g., users’ movie ratings, social network links)? If so, please describe how these relationships are made explicit.}}

There are no relationships between instances that are from distinct geographical locations. However, the label of two different instances originating from the same point I.D  may change over time because of crop rotation. 

\textit{\textbf{Are there recommended data splits (e.g., training, development/validation, testing)? If so, please provide a description of these splits, explaining the rationale behind them.}}

The procured dataset already comes packaged in suggested yearly train/validation/test splits, where instances belonging to one split are not present in another split. Images mapping to the same geographical coordinates are kept together during the splitting process. The dataset is partitioned in such a way  that 70\% of the images of each crop category are reserved for training, 15\% for validation and 15\% for testing. However, it is also possible to combine data from multiple years and recreate unique sets that are favorable to user's needs using the accompanying python toolbox. 

\textit{\textbf{Are there any errors, sources of noise, or redundancies in the dataset? If so, please provide a description.}}

Each image corresponds to a single geographical point taken at varying time periods. Therefore, there is a degree of redundancy in this dataset useful to replicate cases. However, all individual instances are unique. Some images may be affected by small remnants of clouds or discoloration. However, instances with significant noise levels were removed from the cleaned set. 

\textit{\textbf{Is the dataset self-contained, or does it link to or otherwise rely on external resources (e.g., websites, tweets, other datasets)? Please provide descriptions of all external resources and any restrictions associated with them, as well as links or other access points, as appropriate.}}

Yes, the dataset is fully self-contained. There are no restrictions associated with the images, and anyone can access the dataset through the provided \href{https://drive.google.com/drive/folders/1mNI8B5EMk0Xgvx2Pc9ztnQRaW9pXh8yb?usp=sharing}{link}. A \href{https://github.com/bioinfoUQAM/Canadian-cropland-dataset}{GitHub} repository is also provided with benchmarking code and a README.md file to explain how to use and manipulate the dataset.

\textit{\textbf{Does the dataset contain data that might be considered confidential (e.g., data that is protected by legal privilege or by doctor-patient confidentiality, data that includes the content of individuals’ non-public communications)?}}

No, the dataset does not contain any confidential information. All image information is retrieved directly from Google Earth Engine and the Canadian Annual Crop Inventory which are both open-source. 

\textit{\textbf{Does the dataset contain data that, if viewed directly, might be offensive, insulting, threatening, or might otherwise cause anxiety?}} No.

\textit{\textbf{Any other comments?}} N/A.

\subsubsection{Collection Process}

\textit{\textbf{How was the data associated with each instance acquired? Was the data directly observable (e.g., raw text, movie ratings), reported by subjects (e.g., survey responses), or indirectly inferred/derived from other data (e.g., part-of-speech tags, model-based guesses for age or language)? If data was reported by subjects or indirectly inferred/derived from other data, was the data validated/verified? If so, please describe how.}}

A total of 6,633 geographical points corresponding to agricultural fields in Canada were randomly selected using Google Earth Engine. The images were webscraped directly from GEE. The label associated with each field was retrieved using the ACI for that specific year (2017-2020, as of June 9th, 2022). The data was verified using manual observation to exclude any image samples that were deemed low quality. 

\textit{\textbf{What mechanisms or procedures were used to collect the data (e.g., hardware apparatus or sensor, manual human curation, software program, software API)? How were these mechanisms or procedures validated?}}

A python script was created to access the GEE application programming interface (API) and download all the available Sentinel-2 satellite band images per point. If there was no imagery available (due to high cloud or shadow cover post filtering), no imagery was downloaded. 

\textit{\textbf{If the dataset is a sample from a larger set, what was the sampling strategy (e.g., deterministic, probabilistic with specific sampling probabilities)?}}

The dataset represents a fraction of the full ACI. The subset was created by randomly manually tagging points using the 2019 and 2020 ACI, and GEE. Following this, imagery from previous years was extracted from these same points (when available) automatically.  

\textit{\textbf{Who was involved in the data collection process (e.g., students, crowdworkers, contractors) and how were they compensated (e.g., how much were crowdworkers paid)?}}

Two individuals were involved in the data collection process. These include a PhD student and a research scientist at AAFC. The student was paid as a part-time intern while the research assistant was paid as a full-time employee . 

\textit{\textbf{Over what time frame was the data collected? Does this time frame match the creation time frame of the data associated with the instances (e.g., recent crawl of old news articles)? If not, please describe the time frame in which the data associated with the instances was created.}}

The points were collected during the period of mid-march 2021 to mid-april 2021 and January 2022 to March 2022. This time frame does not match the time frame when the data was created (June 2017 - October 2019).  

\textit{\textbf{Were any ethical review processes conducted (e.g., by an institutional review board)? If so, please provide a description of these review processes, including the outcomes, as well as a link or other access points to any supporting documentation.}} No.

\textit{\textbf{Does the dataset relate to people? If not, you may skip the remainder of the questions in this section.}} No. 

\textit{\textbf{Any other comments?}} N/A. 

\subsubsection{Preprocessing / Cleaning / Labeling}

\textit{\textbf{Was any preprocessing/cleaning/labeling of the data done (e.g., discretization or bucketing, tokenization, part-of-speech tagging, SIFT feature extraction, removal of instances, processing of missing values)? If so, please provide a description. If not, you may skip the remainder of the questions in this section.}}

The data was preprocessed by generating sets of images based on the 12 bands retrieved from Sentinel-2. These sets correspond to RGB images, as well as commonly used vegetation indices. These processed images are in .png format and are in their respective sets (RGB, GNDVI, NDVI, NDVI45, OSAVI, PSRI). A color correction algorithm was applied to the RGB dataset. 

Once the images were scraped from the satellite database, a program was created to preprocess the dataset and remove the images with too much cloud, shadows or missing data. This was done using the RGB images. If an image had any cloudy region over the field area, it was automatically removed from the dataset.

\textit{\textbf{Was the “raw” data saved in addition to the preprocessed/cleaned/labeled
data (e.g., to support unanticipated future uses)? If so, please provide a link or other access point to the “raw” data.}}

Yes. The raw dataset for each point is available as a .zip file where all the spectral bands and composite bands are saved in a .tiff format. The raw dataset can be accessed using the GitHub or Google Drive links indicated above. 

\textit{\textbf{Is the software used to preprocess /clean/label the instances available? If so, please provide a link or other access point.}} No.

\textit{\textbf{Any other comments?}} N/A. 

\subsubsection{Uses}
\textit{\textbf{Has the dataset been used for any tasks already? If so, please provide a description.}}

To date, the dataset has been used solely for research tasks. Benchmarking tests were performed in the summer of 2021 at the 42nd Canadian Symposium of Remote Sensing, and for the 2021 NeurIPS Datasets and Benchmarcks Track.    

\textit{\textbf{Is there a repository that links to any or all papers or systems that use the dataset? If so, please provide a link or other access point.}}

No, there is no existing repository for papers or systems using this dataset.

\textit{\textbf{What (other) tasks could the dataset be used for?}}

Following the definitions of the \href{https://arxiv.org/abs/1903.12262}{Montreal Data License}, the dataset can be used for benchmarking (training of models and evaluation of results), research, publishing (models resulting from research) and internal use. The dataset may NOT be used for output commercialization and/or model commercialization. The dataset was created primarily for image classification tasks, but can also be used for data augmentation and the generation of missing images in crop analyses. 

 \textit{\textbf{Is there anything about the composition of the dataset or the way it was collected and preprocessed/cleaned/labeled that might impact future uses?} For example, is there anything that a future user might need to know to avoid uses that could result in unfair treatment of individuals or groups (e.g., stereotyping, quality of service issues) or other undesirable harms (e.g., financial harms, legal risks) If so, please provide a description. Is there anything a future user could do to mitigate these undesirable harms?}

Expansion of the dataset is limited by the availability of Google Earth Engine. However, as it is released now, there are no limitations to the dataset.

 \textit{\textbf{Are there tasks for which the dataset should not be used?} If so, please provide a description.} No. 

 \textit{\textbf{Any other comments?}} N/A.

\subsubsection{Distribution} 
 \textit{\textbf{Will the dataset be distributed to third parties outside of the entity (e.g., company, institution, organization) on behalf of which the dataset was created? If so, please provide a description.}}

Yes, the dataset will be distributed by the University of Quebec in Montreal. It will be distributed to third parties. 

 \textit{\textbf{How will the dataset be distributed (e.g., tarball on website, API, GitHub)? Does the dataset have a digital object identifier (DOI)?}}

The dataset will be distributed via a website hosted at the University of Quebec in Montreal. A GitHub repository with data manipulation tools and machine learning code for the benchmarking tests are available as well. 

 \textit{\textbf{When will the dataset be distributed?}}

The dataset is currently available for download via the links provided in section A.9.2. 

\textit{\textbf{Will the dataset be distributed under a copyright or other intellectual property (IP) license, and/or under applicable terms of use(ToU)? If so, please describe this license and/or ToU, and provide a link or other access point to, or otherwise reproduce, any relevant licensing terms or ToU, as well as any fees associated with these restrictions.}}

The dataset is distributed under the Montreal Data License (see Appendix A.8). 

\textit{\textbf{Have any third parties imposed IP-based or other restrictions on the data associated with the instances? If so, please describe these restrictions, and provide a link or other access point to, or otherwise reproduce, any relevant licensing terms, as well as any fees associated with these restrictions.}} No. 

\textit{\textbf{Do any export controls or other regulatory restrictions apply to the dataset or to individual instances? If so, please describe these restrictions, and provide a link or other access point to, or otherwise reproduce, any supporting documentation.}} No.

\textit{\textbf{Any other comments?}} N/A.

\subsubsection{Maintenance}
\textit{\textbf{Who is supporting/hosting/maintaining the dataset?}}

The University of Quebec in Montreal Bioinformatics laboratory will be hosting the dataset, but it will be continuously maintained by University of Quebec in Montreal and Agriculture and Agrifood Canada.

\textit{\textbf{How can the owner/curator/manager of the dataset be contacted(e.g., email address)?}}

To inquire about this dataset, please email \url{boatswain_jacques.amanda@courrier.uqam.ca}. 

\textit{\textbf{Is there an erratum? If so, please provide a link or other access point.}}

There is no erratum at this point. 

\textit{\textbf{Will the dataset be updated (e.g., to correct labeling errors, add new instances, delete instances)? If so, please describe how often, by whom, and how updates will be communicated to users (e.g., mailing list, GitHub)?}}

Yes, the dataset will be updated at least yearly when the annual crop inventory is available, to retrieve new images and crop labels. There is also the possibility to expand the dataset through the collection of additional points within Canada.  

\newpage

\end{multicols}

\end{document}